\documentclass[preprint,12pt]{elsarticle}

\usepackage{amssymb}
\usepackage{amsmath}
\usepackage{bm}
\usepackage[flushleft]{threeparttable}
\usepackage{multirow}
\usepackage{array}

\begin{document}
\begin{frontmatter}
\title{Learning to count small and clustered objects with application to bacterial colonies} 

\author[kcl-label]{Minghua Zheng\fnref{fn1}}
\author[uh-label]{Na Helian\corref{cor}}
\ead{n.helian@herts.ac.uk}
\author[uh-label]{Peter C. R. Lane}
\author[uh-label]{Yi Sun}
\author[synoptics-label]{Allen Donald}

\cortext[cor]{Corresponding author}
\fntext[fn1]{First author}

\affiliation[kcl-label]{organization={King's College London},
            addressline={St Thomas Street}, 
            city={London},
            postcode={SE1 1UL}, 
            country={United Kingdom}}

\affiliation[uh-label]{organization={University of Hertfordshire},
            addressline={College Lane}, 
            city={Hatfield},
            postcode={AL10 9AB}, 
            country={United Kingdom}}

\affiliation[synoptics-label]{organization={Synoptics Ltd},
            addressline={Beacon House, Nuffield Road}, 
            city={Cambridge},
            postcode={CB4 1TF}, 
            country={United Kingdom}}

\begin{abstract}
Automated bacterial colony counting from images is an important technique to obtain data required for the development of vaccines and antibiotics. However, bacterial colonies present unique machine vision challenges that affect counting, including (1) small physical size, (2) object clustering, (3) high data annotation cost, and (4) limited cross-species generalisation. While FamNet is an established object counting technique effective for clustered objects and costly data annotation, its effectiveness for small colony sizes and cross-species generalisation remains unknown. To address the first three challenges, we propose ACFamNet, an extension of FamNet that handles small and clustered objects using a novel region of interest pooling with alignment and optimised feature engineering. To address all four challenges above, we introduce ACFamNet Pro, which augments ACFamNet with multi-head attention and residual connections, enabling dynamic weighting of objects and improved gradient flow. Experiments show that ACFamNet Pro achieves a mean normalised absolute error (MNAE) of 9.64\% under $5$-fold cross-validation, outperforming ACFamNet and FamNet by 2.23\% and 12.71\%, respectively.
\end{abstract}


\end{frontmatter}

\section{Introduction}
\label{sec-introduction}
Counting small and clustered objects is a special case of generic object counting with important real-world applications. Example applications include bacterial colony counting in medical imaging, bee counting for ecological monitoring, and corn counting in agriculture. While this task is easy for humans, it remains challenging for computer systems due to (1) wide variations in object size, (2) high variations in object density, (3) the lack of labelled data, and (4) various category adaptations.

Previous studies have addressed only some of the aforementioned challenges, but none have tackled them all collectively. For example, the leading cardinality classification method for colony counting~\citep{Ferrari2017} only considers small and clustered colonies but does not address limited labelled data or cross-category adaptations. FamNet~\citep{ranjan_learning_2021} and SAFECount~\citep{you_fewshot_2023} effectively address clustered objects, limited labelled data, and category adaptations, but they overlook small objects. Therefore, the primary goal of this work is to develop algorithms that can learn from limited labelled data, accurately count small and clustered objects, and readily generalise to different object categories.

Bacterial colony counting is used as an example application of the proposed algorithms, with images and annotation software provided by Synoptics Ltd. A \textit{bacterial colony} is a clonal group of cells grown on or within a substance that supports microbial growth~\citep{Jeanson2015}. \textit{Bacterial colony counting} is a process of identifying and enumerating viable bacteria in an image and is widely used in biological laboratories to estimate the number of viable bacteria in a test sample. The counting result is an important indicator of surface cleanliness, product sterility, and bacterial infection. Conventionally, cells are cultured on a circular, transparent, and lidded Petri dish from which a digital camera captures images. The cultured colonies are often small and clustered, making them time-consuming for humans to count and difficult for computer systems. Additionally, colonies from different species vary in colour, shape, opacity, and density, which further
increases the counting difficulty.\footnote{A different colony species corresponds to a different object category.}

This paper poses counting as a few-shot object counting task, treating it as a regression problem based on meta-learning. It also explores adaptations of two existing few-shot object counting methods, namely FamNet and SAFECount, to address small objects with an example application to bacterial colonies. Specifically, the adaptations focus on \textit{feature engineering}. It is a process of formulating the most appropriate features based on data, task, and machine learning algorithm, where a \textit{feature} is a numeric representation of raw data used to train a downstream statistical model~\citep{zheng_feature_2018}. The focus on feature engineering is motivated by the dramatic difference between generic objects and small and clustered objects.

The rest of this paper is organised as follows. Related work about small and clustered object counting is discussed in Section~\ref{sec-related-work}. In Section~\ref{sec-methods}, we introduce data and the proposed algorithms. Experiments and results are presented in Section~\ref{sec-exp}. In Section~\ref{sec-conclusions}, we provide conclusions. Finally, limitations and future work are discussed in Section~\ref{sec-future-work}.

\section{Related work}
\label{sec-related-work}
Previous object counting approaches can be broadly categorised into detection-based, regression-based, density map estimation-based, and few-shot learning-based methods. These approaches are evaluated against the question of whether the counting algorithm can collectively address (1) small object size, (2) object clustering, (3) limited labelled data, and (4) category adaptation. This paper explores applications in generic object counting and bacterial colony counting with a primary focus on the latter because it is the only dataset used in this study. Counting methods that rely on large language models or hardware design are not considered, as they fall outside the scope of this work.

\subsection{Detection-based counting}
A common way to count objects is the detect-then-count approach, which consists of two steps. The first step is to detect all instances of the target object from an image. The final count is then obtained by summing the detections. The first step is a challenging task in itself that has been studied in computer vision for many years. Detection methods can be further categorised into traditional image processing approaches, machine learning approaches, and hybrid approaches that combine both.

\subsubsection{Counting by traditional image processing approaches}
Traditional image processing approaches detect objects by identifying boundaries or regions. Boundary detection can be achieved using edge detection~\citep{Barber2001, Loukas2003, Choudhry2016}, contour detection~\citep{Niyazi2007, Wienert2012}, and Hough Transform~\citep{DanaH.Ballard1981, Flaccavento2011, Geissmann2013, Matic2016}. Region detection can employ connected components labelling~\citep{Geissmann2013}, thresholding~\citep{Zhang2007, Ates2009, Chen2009, Smith2014, Chiang2015, Khan2018}, and template matching~\citep{Kachouie2009}. After identifying boundaries or regions, fixed parameters or separability criteria are often applied to distinguish clustered objects and reject outliers, either pre-defined or specified at runtime. A special region detection approach which combines distance transform and watershed algorithm has also been used to assess colony separability before splitting for counting~\citep{Clarke2010,wong_republication_2019}.

The main limitation of traditional image processing approaches is their reliance on user-specified parameters, which makes automatic adaptation to varying object sizes, densities, and categories difficult. For example, a colony cluster is kept for counting if its area is greater than 30\% of the median colony area, which may not generalise to a different colony species~\citep{Khan2018}. Similarly, criteria for identifying clustered colonies are hard to automate because of human intervention and inflexible parameters. For instance, users must specify the minimal radius, area, and height before detecting clustered colonies~\citep{Geissmann2013}. Additionally, region detection approaches that incorporate the watershed algorithm may become less accurate due to over-segmentation~\citep{Qin2013}. Despite these limitations, such approaches do not require annotated training data and can adapt to new categories through manual parameter adjustment.

\subsubsection{Counting by machine learning approaches}
\label{chapter2_sub_sec_ml_approaches}
Machine learning algorithms can overcome the inflexibility of traditional image processing methods because they can learn patterns from data to perform specific tasks without being explicitly programmed. However, developing such algorithms typically requires extensive data collection and annotation. Category adaptation is also challenging since most machine learning algorithms are dependent on training data and require retraining to generalise to new categories. Moreover, little is currently known about how effectively machine learning algorithms address small and clustered objects.

Many machine learning algorithms for object counting have paid little attention to clustered objects, often producing inaccurate counts. For example, K-means clustering was used to group pixels into background, colonies, and artefacts, followed by manual selection to train a support vector machine for species identification~\citep{Chen2009}. However,~\citet{Chen2009} overlooked clustered colonies and introduced a hurdle for full automation. Similarly, segmentation algorithms fail to distinguish clustered objects, such as convolutional neural networks (CNNs)~\citep{Andreini2018, Hilsenbeck2017, Liu2017_, Sadanandan2017}, U-Net~\citep{ronneberger_unet_2015}, and U-Net-based models~\citep{falk_unet_2019, zhou_unet_2018, cao_u2net_2024}. Although Mask R-CNN~\citep{He2017} has been adapted for colony detection and species classification, it was evaluated on a dataset with few clustered colonies and two species. Another neural network classified pixels into the border of clustered colonies, background, virulent colony or avirulent colony~\citep{Beznik2020}, but the outcome can only reveal colony area rather than the actually colony count.

Generic object detection algorithms often use neural networks to predict bounding boxes and corresponding objectness scores for target objects, thereby localising objects within different regions of an image. Popular approaches include R-CNN-based models, such as R-CNN~\citep{Girshick2014, majchrowska_assessing_2025}, Fast R-CNN~\citep{Girshick2015}, and Faster R-CNN~\citep{Ren2017}. Another major family of methods comprises YOLO-based models, including YOLO \citep{Redmon2015}, YOLO9000 \citep{Redmon2017}, YOLOv3 \citep{Redmon2018}, YOLOv4 \citep{bochkovskiy_yolov4_2020}, YOLOv5 \citep{Jocher_ultralytics_2022}, YOLOv6 \citep{li_yolov6_2022}, and YOLOv7 \citep{wang_yolov7_2022}. Although widely applied across domains, none of these methods explicitly addresses clustered objects. Moreover, they require time-consuming bounding box annotations for training.

To address the difficulty of handling clustered colonies, some researchers employed a CNN to classify each colony cluster into a pre-defined cardinality class~\citep{Ferrari2015, Ferrari2017}. The number of colonies in a plate image is obtained by summing the predicted cardinalities of all detected clusters. Nevertheless, this method is dependent on the detection of each colony cluster. The impact of high visual similarity among colony clusters on counting performance remains unknown, and class imbalance is not considered.

\subsubsection{Counting by hybrid approaches}
Due to inflexible pre-defined parameters, the need for user intervention, and difficulty with clustered objects, researchers have combined image processing methods and machine learning algorithms to count bacterial colonies. For example, \citet{Brugger2012} combined thresholding with Bayes classification to detection colonies. However, circular colonies were assumed in their work, which could be problematic for colonies of a different shape. Despite the hybrid design, the challenges of counting of small and clustered objects, the lack of labelled data, and the need for category adaptation remain unsolved.

\subsection{Regression-based counting}
Regression-based counting approaches avoid object detection by directly predicting object counts from images. These approaches have been used in crowd counting since the spatial information and the distribution of people are not needed for the final count. For example, hand-coded features~\citep{davies_crowd_1995} and twenty-nine perspective-normalised features~\citep{chan_bayesian_2009} were used to build regression models for crowd counting. Nevertheless, regression-based approaches are only suitable when the object localisation and spatial distribution are not essential to the application. This limitation restricts their adaptability to other category adaptations, where spatial information is often important. It is also unknown what contributes to the count since the output is merely a number. Like other learning-based algorithms, regression-based counting approaches still require substantial labelled data.

\subsection{Density map estimation-based counting}
Density map estimation-based counting approaches avoid human intervention and the hard task of object detection. Approaches in this category learn to map an input image to a density map in which the sum of density values represents the object count. Density map estimation is particularly useful to count clustered objects because the predicted density value can range from 0 to any positive number. Additionally, the predicted density map can represent a general location of objects. However, the main drawback is the requirement of density map annotations, even though they are significantly easier than bounding box annotations to create. Similar to regression approaches, algorithms must be retrained to adapt to a different category.

Despite the two aforementioned drawbacks, density map estimation has been widely applied to crowd counting. One possible reason is that category adaptation is less critical in crowd counting. According to the survey by~\citet{fan_survey_2022}, density map-based crowd counting approaches can be grouped into several categories. They include multi-scale~\citep{boominathan_crowdnet_2016}, context-aware~\citep{amirgholipour_accnn_2018}, auxiliary-task~\citep{huang_body_2018}, limited-label, domain-adaptation, perspective-guided~\citep{yan_perspectiveguided_2019}, attention-based~\citep{liu_adcrowdnet_2019}, and network architecture search models~\citep{hu_nascount_2020}. Among these categories, only limited-label models that address limited labelled data and domain-adaptation models are reviewed here, as they are more relevant to the main aim of this work.

Limited-label and domain-adaptation models for crowd counting are often based on Generative Adversarial Networks (GANs). This is because GANs can generate synthetic images or features that closely align with the distribution of real data~\citep{goodfellow_generative_2020}. For example,~\citet{olmschenk_dense_2019a} modified the traditional discriminator in GANs to produce a value of the expected crowd count and a flag to indicate if the input is real. The modified discriminator is more effective to capture features from data because it is a result of combining supervised regression and unsupervised classification. Similarly,~\citet{wang_learning_2019a} proposed a model called SSIM Embedding Cycle GAN to adapt to a different domain by translating synthetic crowd scenes to real scenes. Nevertheless, GAN-based models are well known for their training instability and optimisation difficulty~\citep{saxena_generative_2021}.

Inspired by advances in crowd counting, researchers have been using density map estimation to count cells and colonies with promising results~\citep{albaradei_automated_2020, xie_microscopy_2018, xue_cell_2016, graczyk_selfnormalized_2022}. However, these approaches still suffer from the two aforementioned drawbacks.

\subsection{Few-shot learning-based counting}
To address the lack of labelled data and category adaptation in object counting, some researchers~\citep{ranjan_learning_2021, you_fewshot_2023} have explored few-shot object counting. The goal is to count the number of exemplar objects presented in a query image where exemplar objects are described in only a few labelled support images.\footnote{The statement that few-shot object counting approaches address the lack of labelled data means these approaches only use \bf{a few labelled support images}.} Object classes are divided into base classes for training and novel classes for testing, where the base classes have no intersection with the novel classes. During training, the model learns from the query image and a few support images with ground truth density map. During testing, the model predicts a density map for a given query image with only a few support images by leveraging the knowledge gained from the base classes. These few-shot counting approaches have demonstrated promising results on generic objects. Nevertheless, their effectiveness for small objects remains largely unexplored.

Two few-shot counting models have shown great potential to address clustered objects, limited labelled data, and category adaptation. FamNet uses a pre-trained ResNet-50~\citep{He2015} to extract features from query and support images, followed by a feature correlation layer and a regression module to predict the final density map~\citep{ranjan_learning_2021}. Category adaptation is achieved by using exemplars and a newly proposed adaptation loss function to fine tune the regression module. Despite its promising results, FamNet does not explicitly address small objects and is not end-to-end trainable, leaving the model sub-optimised for the counting task. Its domain adaptation also requires test time training, which may not satisfy a busy laboratory. SAFECount (\underline{S}imilarity-\underline{A}ware \underline{F}eature \underline{E}nhancement block for object \underline{Count}ing) integrates a transformer-based attention mechanism~\citep{vaswani_attention_2017} to produce a clearer boundary between objects in the predicted density map~\citep{you_fewshot_2023}. While SAFECount has yet to be applied to count small objects, it no longer requires test time training to adapt to a different domain due to the attention mechanism.

\section{Methods}
\label{sec-methods}
\subsection{Data}
\label{subsec-data}
\begin{figure}[h!]
\centering
\includegraphics[width=.40\linewidth]{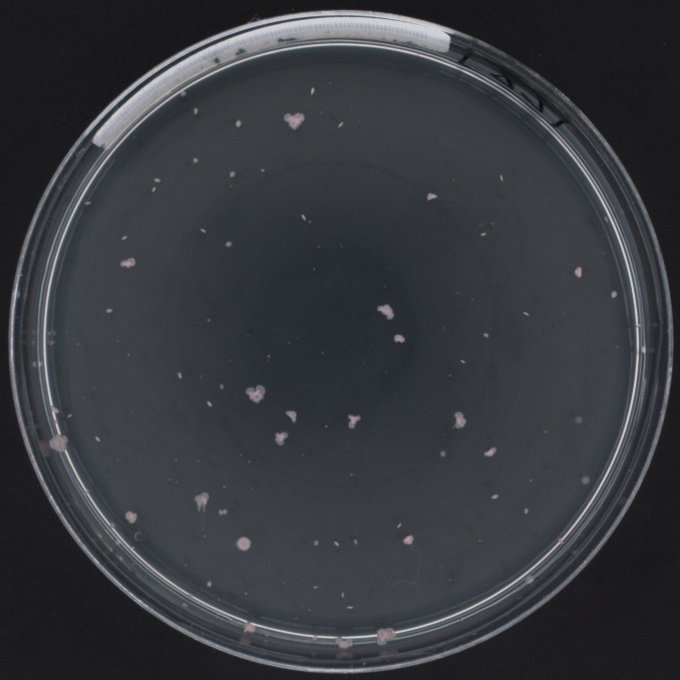}
\includegraphics[width=.40\linewidth]{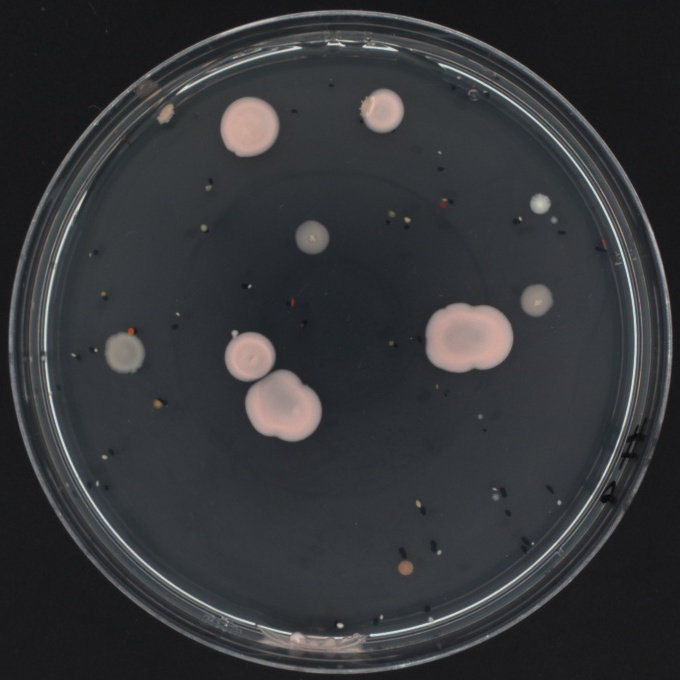}
\includegraphics[width=.40\linewidth]{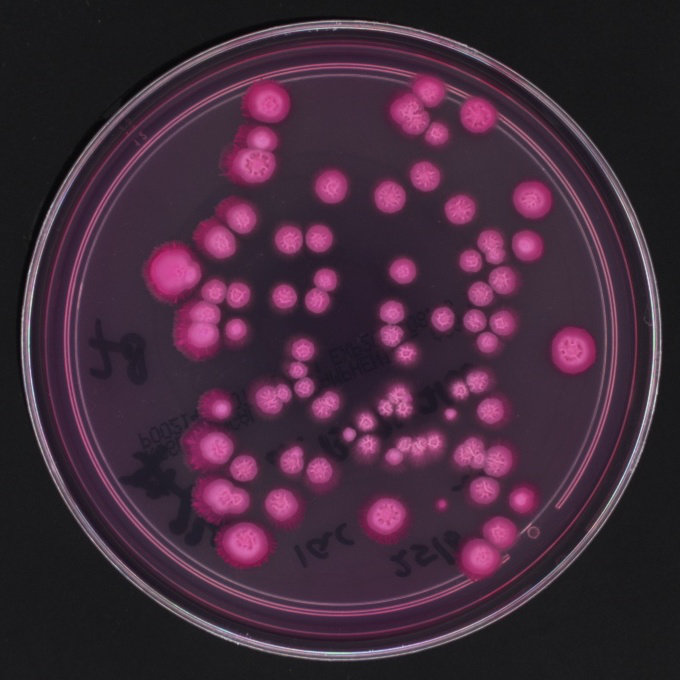}
\includegraphics[width=.40\linewidth]{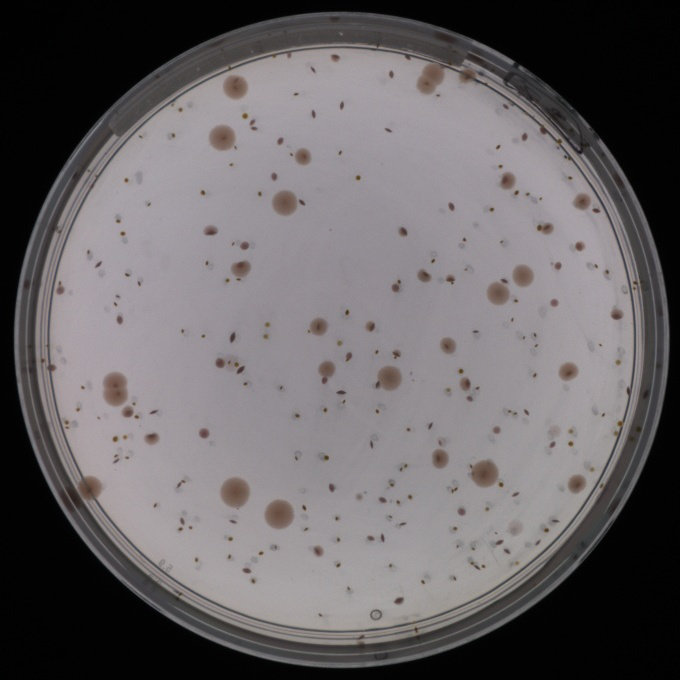}
\caption{Four example plate images with colonies of different species, sizes, and colours.}
\label{example-plate-img1}
\end{figure}

A collection of real-world plate images of colonies was provided by Synoptics Ltd to construct the Synoptics Dataset. The original plate images are cropped in Photoshop from~3$\times$1040$\times$1040~to~3$\times$680$\times$680~to remove non-Petri-dish areas~\citep{zheng_bacterial_2022}, resulting in 125 unique images. Although all images have the same format and Petri-dish shape, the colony species are unknown and vary across the dataset. Some example plate images with colonies of
different species, colours, sizes, and shapes are illustrated in Figure~\ref{example-plate-img1}.

\paragraph{Image annotation}
The colony centres and bounding boxes are created with the assistance of proprietary software, ProtoCOL, provided by Synoptics Ltd.\footnote{https://www.synbiosis.com/product/automated-colony-counting-zone-measurement-protocol-3} ProtoCOL requires careful and manual parameter adjustment for single colony and clustered colonies in each image, making the process time-consuming. Additionally, colony centres and bounding boxes must be manually checked and corrected by the first author before using them for subsequent analyses.

\paragraph{Data splitting}
\label{subsec_dataset_split}
The Synoptics Dataset is randomly split into a fixed training set and test set with a ratio of 8:2. This ratio ensures that 20\% of the whole data is kept as a test set, as suggested by~\citet{ranjan_learning_2021}. Therefore, the training set and test set consist of 100 and 25 images, respectively. Because different images in this dataset may have the same colony species, it is assumed that images in the training set contain the same colony species in the test set. This implies the assumption that base classes do not overlap with novel classes in few-shot object counting~\citep{you_fewshot_2023} is broken when this dataset is used to develop few-shot object counting algorithms.

\begin{figure}[h!]
\centering
\includegraphics[width=.75\linewidth]{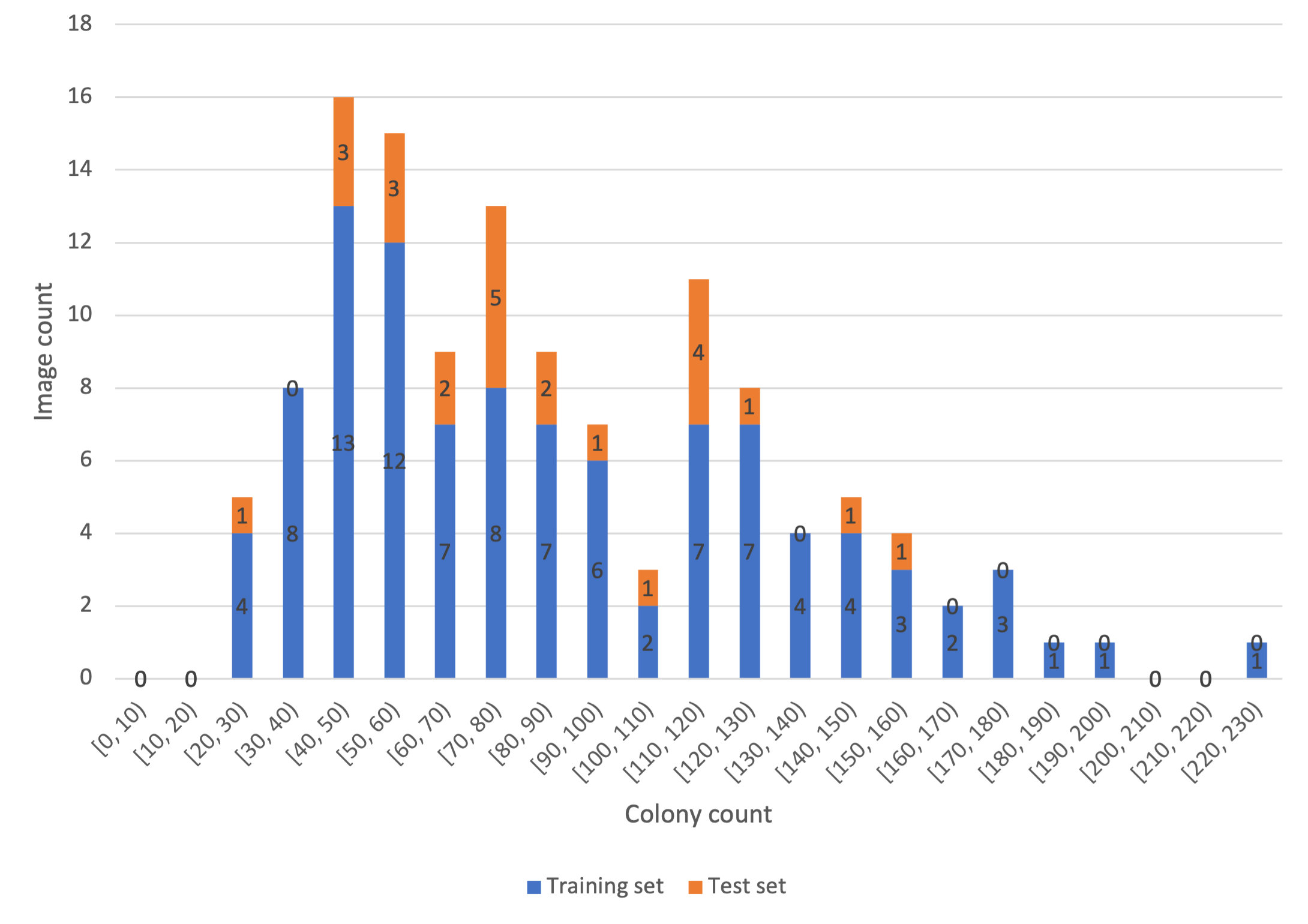}
\caption{Distribution of colony counts in the training and test sets.}
\label{fig:stat_train_test}
\end{figure}

\paragraph{Data statistics}
Figure~\ref{fig:stat_train_test} illustrates the distribution of colony counts across different sets, where the $x$ axis represents colony counts grouped in intervals of 10 and the $y$ axis shows the number of plate images within each interval. Overall, each plate image contains at least 20 colonies. Colony counts between 40 and 50 are more frequent in the training set and the whole set, whereas colony counts between 70 and 80 occur more often in the test set. In addition to this distribution analysis, the mean and variance of pixel values for training set, test set, and the whole set are summarised in Table~\ref{tab_data_v2_mean_std_var}. The values in each set of parentheses correspond to the values calculated from the red channel, green channel, and blue channel across all images in the dataset, where the pixel value ranges from 0 to 255. The mean pixel values of the training and test sets are similar, but the test set has a lower variance in pixel values than the training set.

\begin{table*}[t]
\centering
\caption{Mean and variance of pixel values in the training set, test set, and the whole set.}
\label{tab_data_v2_mean_std_var}
\begin{tabular}{lcc}
        \hline
        Data      & Mean                  & Variance                 \\ \hline
        Training  & (54.43, 56.19, 60.98) & (903.25, 884.39, 927.27) \\
        Test      & (53.48, 55.44, 60.03) & (820.91, 819.79, 861.32) \\
        The whole & (54.24, 56.04, 60.79) & (886.93, 871.56, 914.22) \\ \hline
\end{tabular}
\end{table*}

\subsection{ACFamNet}
\label{subsec-acfamnet}
\subsubsection{Overview}
\begin{figure}[h!]
\centering
\includegraphics[width=.75\linewidth]{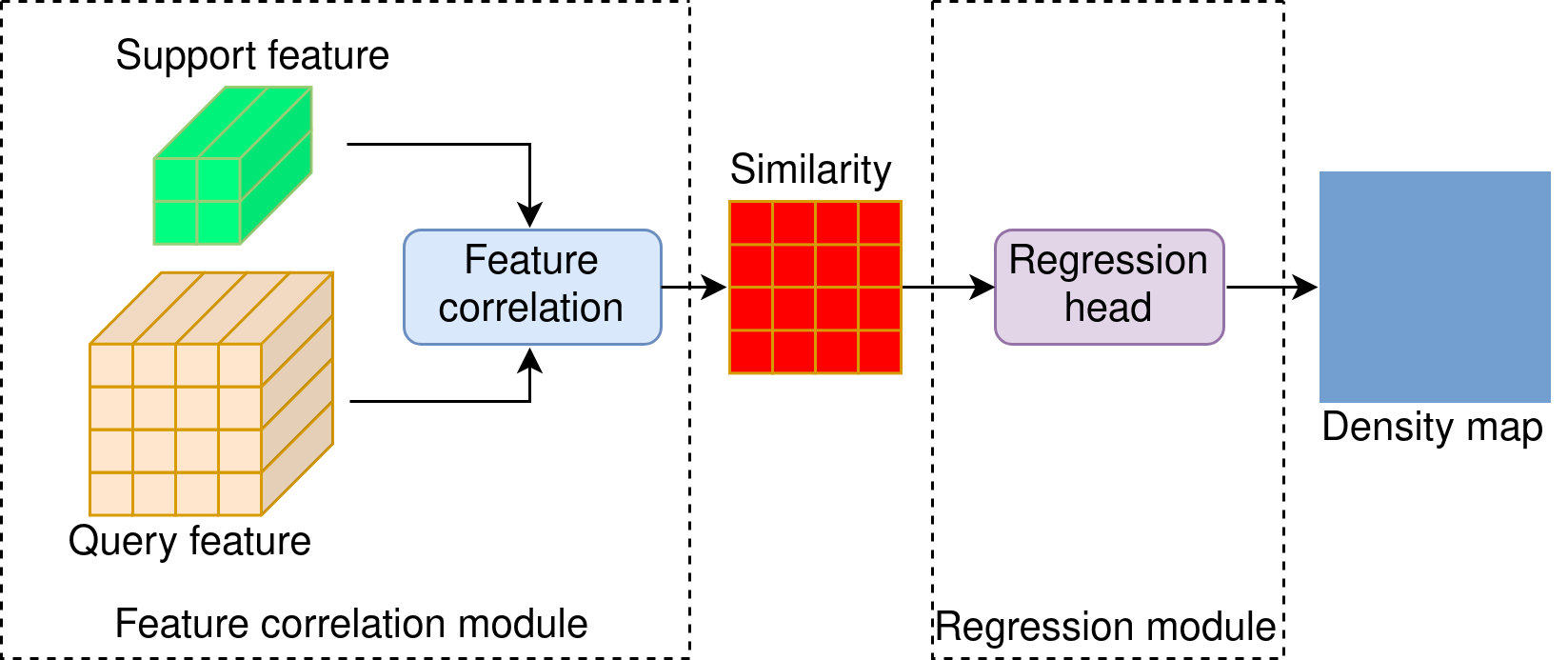}
\caption{Overall structure of ACFamNet. The feature correlation and regression modules are detailed in Figure~\ref{fig:acfamnet_feature_correlation_module} and Figure~\ref{fig:acfamnet_regression_module_overview}.}
\label{fig:acfamnet_concept_overview}
\end{figure}

The overall structure of the \underline{A}ligned \underline{C}ustom \underline{F}ew-shot \underline{A}daptation and \underline{M}atching \underline{Net}work (ACFamNet) is illustrated in Figure~\ref{fig:acfamnet_concept_overview}. It is an end-to-end trainable model with two modules: a feature correlation module and a regression module. In the former module, feature correlation is performed between support and query features, derived from the exemplars and the input image, respectively. The output of feature correlation is a similarity map that is fed into the regression module to produce the final density map.

\subsubsection{Feature correlation module}
The feature correlation module in ACFamNet consists of a simple convolutional layer with $k$ kernels of size~7$\times$7~to extract features from the input image, where $k$ is a hyper-parameter. This convolution is performed by a stride of 2 with 0 padding added to preserve the input image's spatial information, followed by batch normalisation~\citep{ioffe_batch_2015} and ReLU activation~\citep{agarap_deep_2018}. The output is referred to as query feature. The kernel, stride, and padding sizes are identical to those of the first convolutional layer in ResNet-50~\citep{He2015}, which is widely used in neural network architectures.

\begin{figure}[h!]
\centering
\includegraphics[width=1\linewidth]{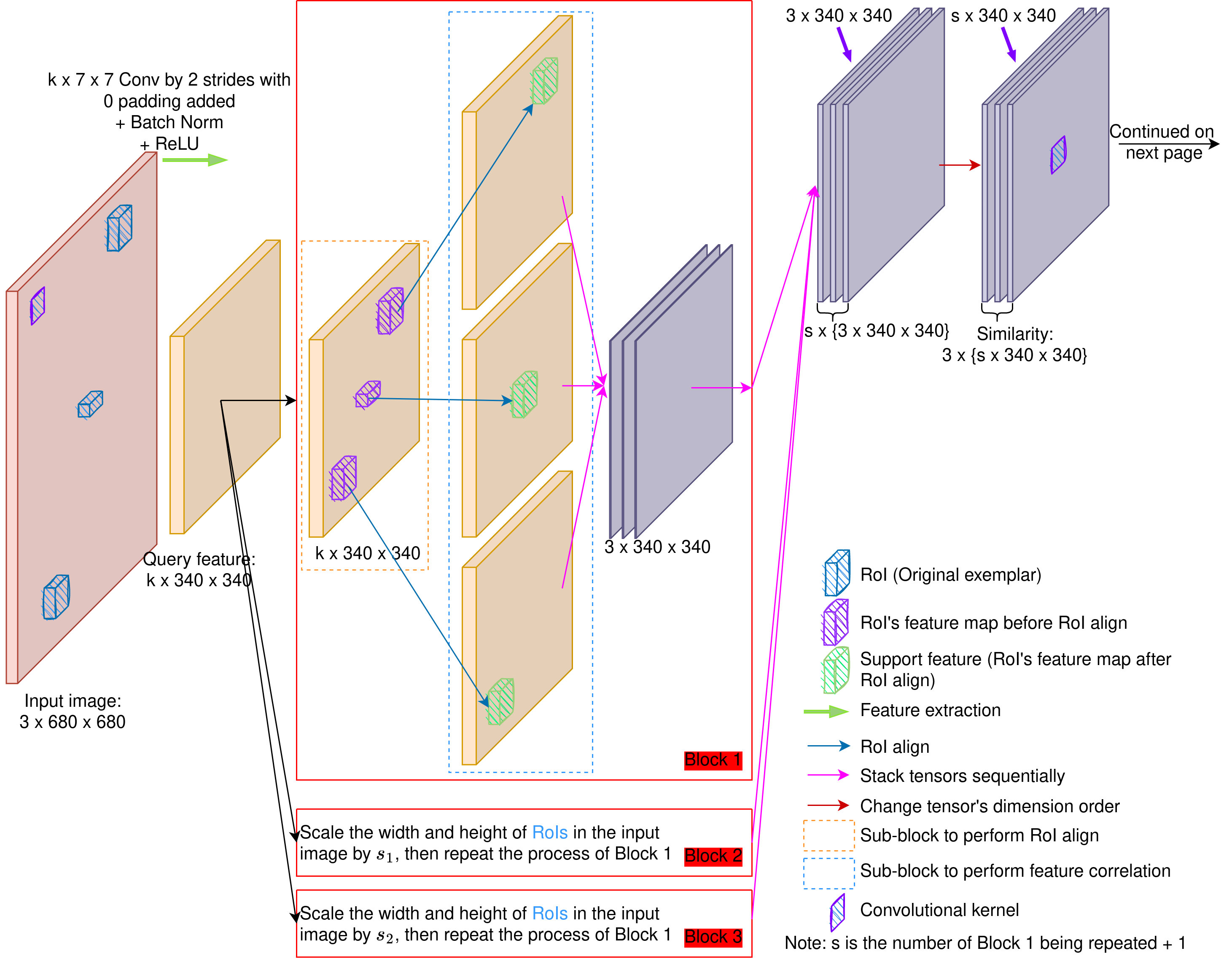}
\caption{Illustration of the ACFamNet feature correlation module.}
\label{fig:acfamnet_feature_correlation_module}
\end{figure}

Let the $i$-th of $K$ exemplars (blue cubes in Figure~\ref{fig:acfamnet_feature_correlation_module}), which is also referred to as Region of Interest (RoI), be defined in the input image of size $H \times W$ by a bounding box $\bm{b}_i = (x_i, y_i, h_i, w_i)$. With the query feature map (orange cube in Figure~\ref{fig:acfamnet_feature_correlation_module}) downsampled by 2, the exemplar coordinates are proportionally projected onto the feature space as $\frac{1}{2}\bm{b}_i$ without coordinate quantisation. This projection corresponds to the purple cubes in Figure~\ref{fig:acfamnet_feature_correlation_module}. Given the query feature map $\bm{f_Q} \in \mathbb{R}^{k \times H_Q \times W_Q}$, RoI Align~\citep{He2017} is applied to the projected region to extract a fixed-size support feature of $H_S \times W_S$ (green cubes in Figure~\ref{fig:acfamnet_feature_correlation_module}):

\begin{equation} \label{acfamnet_RoIAlign}
    \bm{f_{Si}} = \text{RoIAlign}(\bm{f_Q}, \frac{1}{2}\bm{b}_i), \quad \bm{f_{Si}} \in \mathbb{R}^{k \times H_S \times W_S}
\end{equation}

The support feature is then treated as a convolutional kernel and correlated with the query feature map to generate a similarity map. This operation, which is illustrated as Block 1 in Figure~\ref{fig:acfamnet_feature_correlation_module}, measures spatial similarity between the exemplar and all locations in the query feature map. The similarity maps from all $K$ exemplars are concatenated along the channel dimension, resulting in $\bm{S} \in \mathbb{R}^{K \times H_Q \times W_Q}$:

\begin{equation} \label{acfamnet_feature_correlation} 
    \bm{S} = \text{concat}_{i=1}^{K}(\text{conv}(\bm{f_Q}, \text{kernel})), \quad \text{kernel} = \bm{f_{Si}}
\end{equation}

The feature correlation process (Block 1 in Figure~\ref{fig:acfamnet_feature_correlation_module}) is repeated another two times with the original exemplars scaled by two different factors ($s_1$ and $s_2$ in Figure~\ref{fig:acfamnet_feature_correlation_module}), aiming to tackle the same exemplar of different sizes. Similarly, the outputs are stacked together before reorganising the dimension order based on the exemplar's dimension. This is because ACFamNet prioritises features from each exemplar over features from the resized version. It will be showing that ACFamNet will suffice without repeating feature correlation multiple times in the experimental section.

\subsubsection{Regression module}
\begin{figure}[h!]
\centering
\includegraphics[width=.75\linewidth]{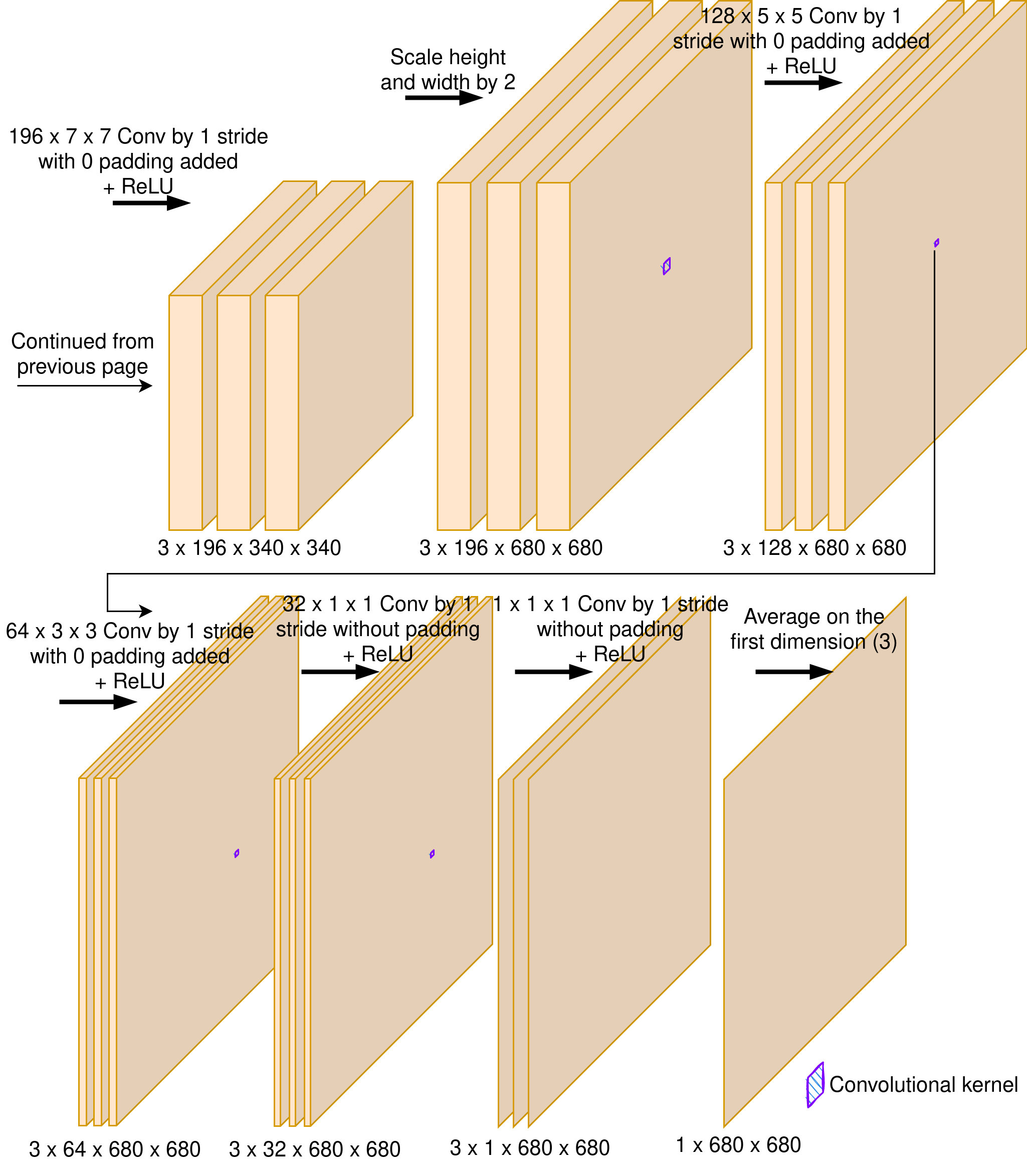}
\caption{Illustration of ACFamNet regression module.}
\label{fig:acfamnet_regression_module_overview}
\end{figure}

Similar to FamNet's density prediction module, the ACFamNet regression module receives the similarity map from the previous module as input to predict a density map. As illustrated in Figure~\ref{fig:acfamnet_regression_module_overview}, the regression module consists of five convolutional layers and an upsampling layers placed after the first convolutional layer. The upsampling layer doubles the input's height and width using bilinear interpolation algorithm. The kernel parameters, which are the number of kernels, height and width, for these five convolutional layers are~196$\times$7$\times$7,~128$\times$5$\times$5,~64$\times$3$\times$3,~32$\times$1$\times$1, and~1$\times$1$\times$1, respectively. The first, second, and third convolutional layers have 0 padding added to preserve the input's height and width. All convolutions are performed by a stride of 1, and ReLU activates the output of each convolutional layer. Finally, the output is averaged along the exemplar dimension to produce a one-dimensional density map with the same height and width as the input image.

\subsubsection{Comparison with FamNet}
ACFamNet is heavily inspired by FamNet, which can effectively count 147 types of generic objects using only three exemplars that act as templates to locate corresponding instances in the input image. This few-shot learning-based strategy tackles the difficulty of collecting and annotating data by only using these exemplars. Compared with FamNet, ACFamNet is fully end-to-end trainable, enabling all modules to be differentiable and jointly optimised for the task. The end-to-end trainable model has been proven effective in complex tasks such as speech recognition~\citep{collobert_natural_2011, lewis_deal_2017}, machine translation~\citep{wu_googles_2016}, and autonomous driving~\citep{serban_standard_2018}. Additionally, ACFamNet replaces RoI pooling with RoI Align to mitigate the RoI misalignment for small objects, where RoI pooling may cause significant information loss. Moreover, ACFamNet has a simplified feature extraction module consisting of a single convolutional layer, substantially reducing computational cost without degrading performance. Finally, ACFamNet requires only a single scale factor, compared to the three scale factors in FamNet, which further improves computational efficiency.

\subsection{ACFamNet Pro}
\label{subsec-acfamnetpro}
ACFamNet Pro is an advanced version of ACFamNet with an additional multi-head attention mechanism and residual connections. This design is inspired and motivated by SAFECount, published during the development of ACFamNet, and ResNet-50. Specifically, the multi-head attention mechanism in SAFECount dynamically weights objects of interest, producing better feature representations. The residual connections in ResNet-50 help optimise gradient flow, thereby improving task performance.

\subsubsection{Overview}
\begin{figure}[h!]
\centering
\includegraphics[width=.85\linewidth]{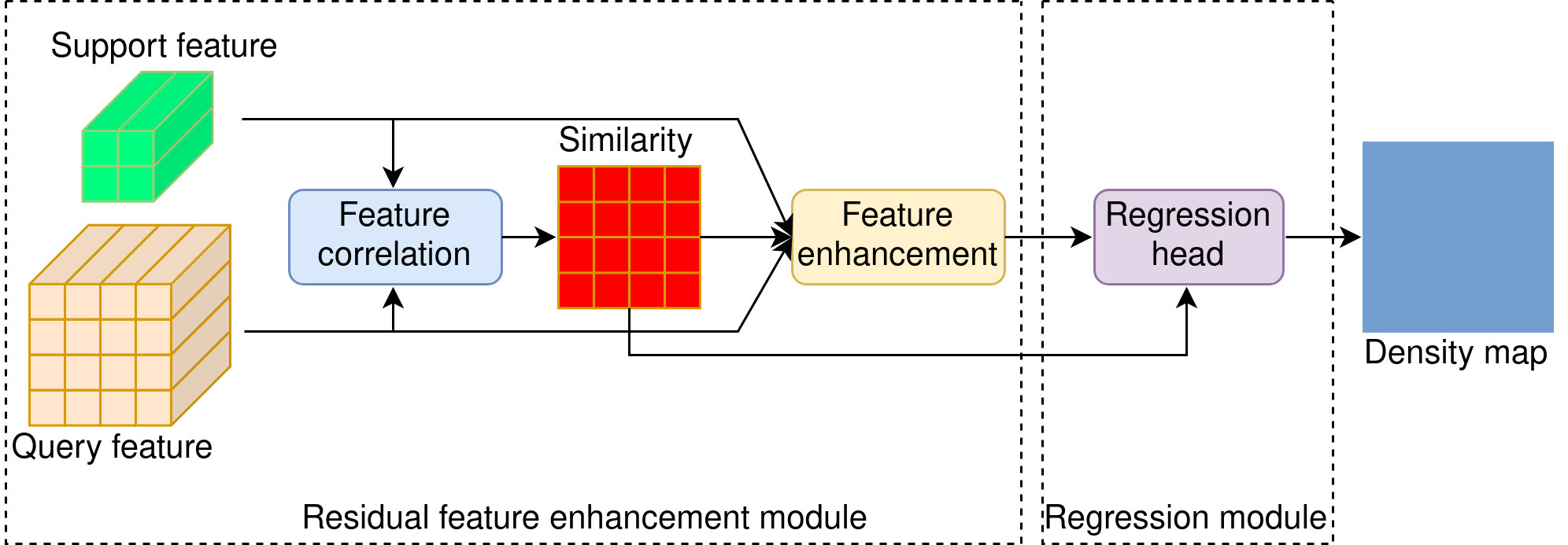}
\caption{Overall structure of ACFamNet Pro. Details of the residual feature enhancement module and the regression module are provided in Figure~\ref{fig:acfamnetpro_residual_fe_module} and Figure~\ref{fig:acfamnetpro_regression_module}, respectively.}
\label{fig:acfamnetpro_concept_overview}
\end{figure}

The overall structure of ACFamNet Pro is illustrated in Figure~\ref{fig:acfamnetpro_concept_overview}. It is an end-to-end trainable model composed of a residual feature enhancement module and a regression module. In the former module, feature correlation is performed between support and query features, which are derived from the exemplars and input image, respectively. The output of feature correlation is a similarity map which is fused with the support feature and query feature to enhance features. The enhanced features along with the similarity map are fed into the regression module to produce the final density map.

\subsubsection{Query and support features}
\begin{figure}[h!]
\centering
\includegraphics[width=.85\linewidth]{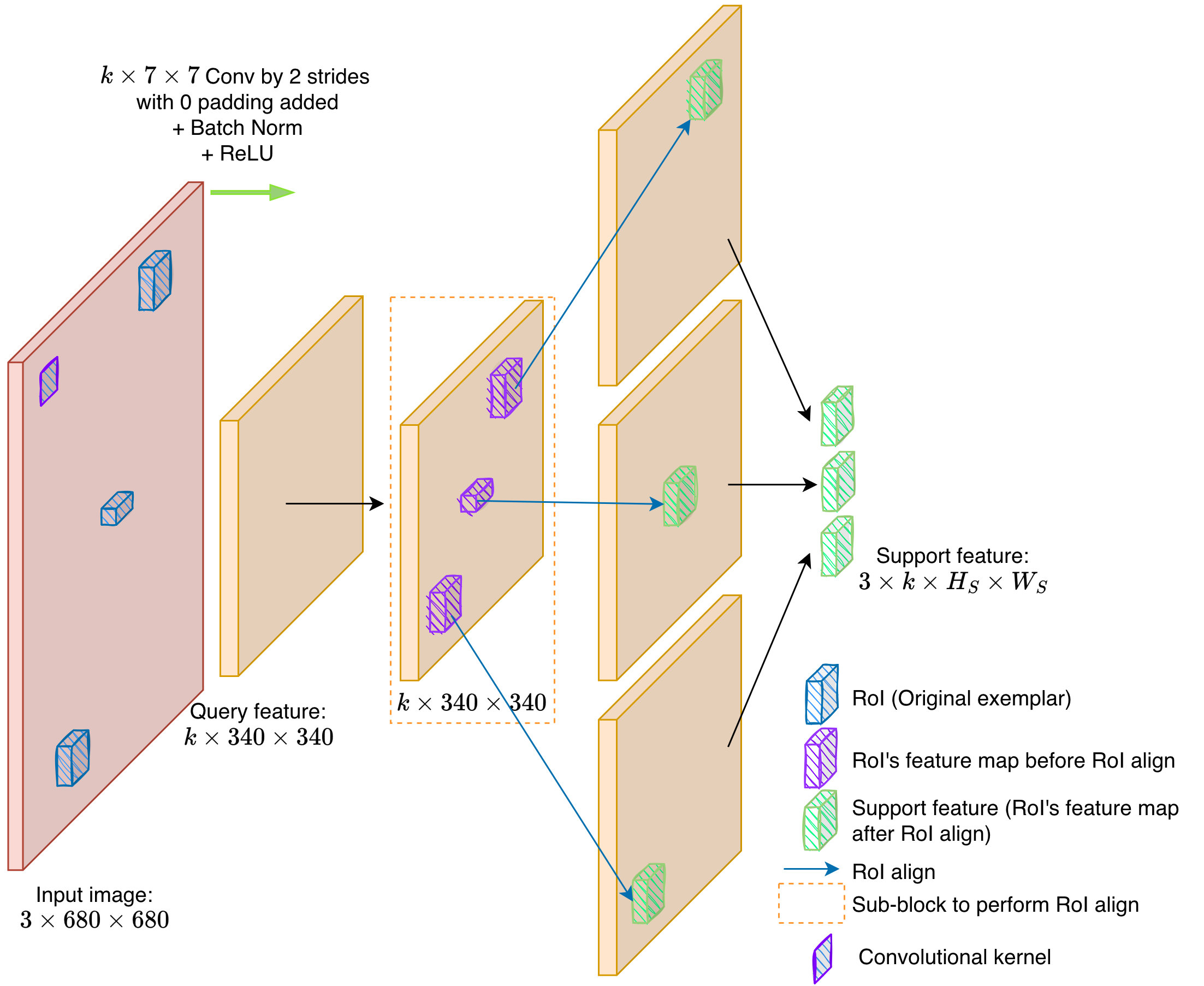}
\caption{Feature extractor.}
\label{fig:acfamnetpro_feature_extractor}
\end{figure}

The query and support features are extracted by a feature extractor illustrated in Figure~\ref{fig:acfamnetpro_feature_extractor}. This feature extractor is often known as \emph{backbone} in literature~\citep{tan_efficientdet_2020, qin_u2net_2020}. Specifically, the feature extractor consists of a single convolutional layer with $k$ kernels of size~7$\times$7, where $k$ is a hyper-parameter. The convolution is performed by a stride of 2 with 0 padding added to preserve the input image's spatial information, followed by batch normalisation and ReLU activation. This feature extractor is identical to the first convolutional layer of ACFamNet shown in Figure~\ref{fig:acfamnet_feature_correlation_module}. The resulting query feature is denoted as $\bm{f_Q} \in \mathbb{R}^{k \times H_Q \times W_Q}$, where $H_Q$ and $W_Q$ are equal to half the height and width of the input image, respectively.

The support image is commonly cropped from the query image to include only the specified exemplar. Therefore, the support feature can be obtained by applying RoI Align to the query feature. This process is repeated $K$ times if $K$ support images are used. The resulting support feature is denoted as $\bm{f_S} \in \mathbb{R}^{K \times k \times H_S \times W_S}$, where $H_S$ and $W_S$ are the height and width of the RoI Align output, respectively.

\subsubsection{Residual feature enhancement module}
The design of residual feature enhancement module mimics the multi-head attention mechanism used in transformers. This is because this mechanism has demonstrated strong effectiveness in many natural language process applications~\citep{vaswani_attention_2017} and computer vision applications~\citep{dosovitskiy_image_2020}. This mechanism describes a weighted average of elements with the weights dynamically calculated based on an input query and elements' keys, where the element is interpreted as the support feature in ACFamNet Pro. Additionally, this mechanism allows the model to control the mixing of information between elements, i.e. support feature, to enrich feature representations.

\begin{figure}[h!]
\centering
\includegraphics[width=1\linewidth]{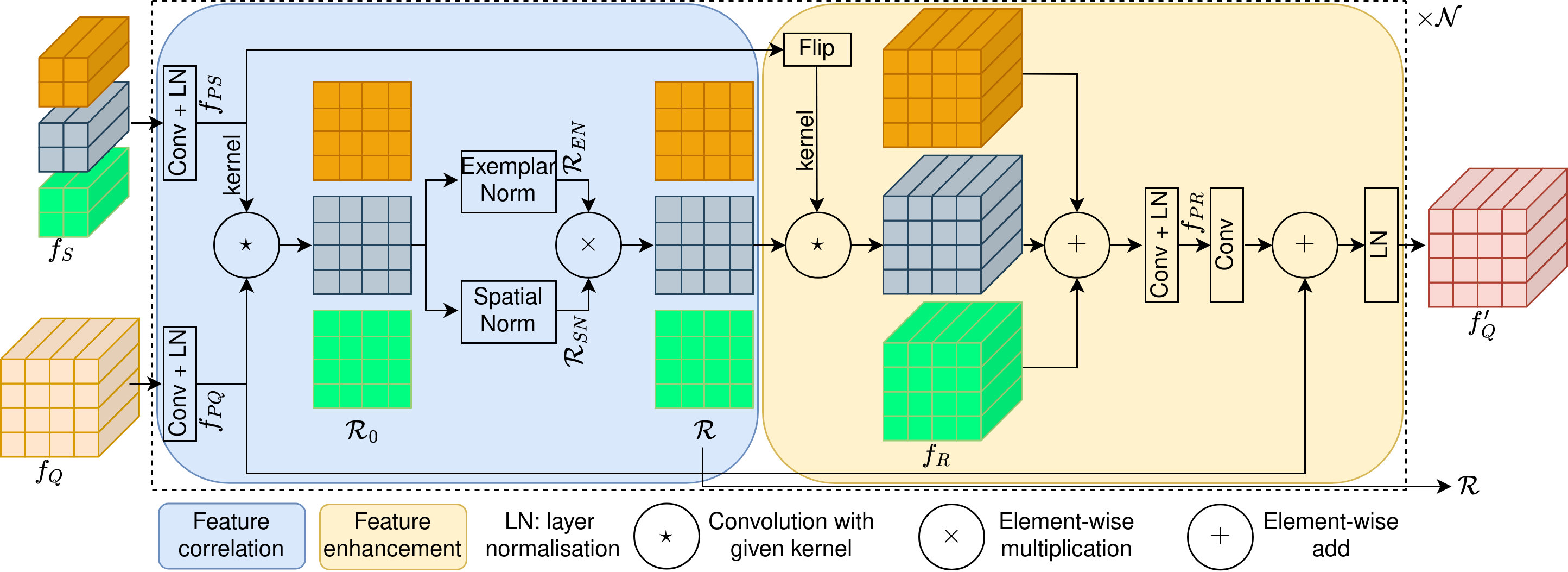}
\caption{Residual feature enhancement module.}
\label{fig:acfamnetpro_residual_fe_module}
\end{figure}

The detailed design of the residual feature enhancement module is illustrated in Figure~\ref{fig:acfamnetpro_residual_fe_module}. It firstly projects the query feature and support feature into the same feature space, followed by a comparison at every spatial position to produce a score map. In the feature correlation block, the multiple score maps generated by the support images are concatenated, and then normalised along both the exemplar and spatial dimensions to produce a reliable similarity map. In the feature enhancement block, the similarity map is used as weights to integrate the support feature into the query feature to produce an enhanced feature. The enhanced feature along with the similarity map are fed into a regression module to produce a density map, where the similarity map is used as a residual connection to improve the regression module. Finally, the whole residual feature enhancement module can be stacked multiple times to further enhance feature representations.

\paragraph{Feature correlation block}
\label{subsubsec_feature_correlation}
The aim of feature correlation block is to produce a similarity map to robustly highlight regions in the query feature $\bm{f_Q}$ that are similar to the support feature $\bm{f_S}$. It has three steps: learnable feature projection, feature comparison, and score normalisation.

\subparagraph{Learnable feature projection}
The useful features from the support feature $\bm{f_S}$ and query feature $\bm{f_Q}$ are dynamically selected by~1$\times$1~convolution with $C$ kernels. Another purpose of this convolution is that both support feature $\bm{f_S}$ and query feature $\bm{f_Q}$ are projected into the same feature space so that they can be compared. The convolution is followed by a layer normalisation to bring these two features to the same distribution. The outputs are referred to as \emph{projected support feature} and \emph{projected query feature} with updated notations: $\bm{f_{PS}} \in \mathbb{R}^{K \times C \times H_S \times W_S}$ and $\bm{f_{PQ}} \in \mathbb{R}^{C \times H_Q \times W_Q}$. In practice, $\bm{f_S}$ and $\bm{f_Q}$ share the same convolution layer and layer normalisation layer because $\bm{f_S}$ is cropped from $\bm{f_Q}$ followed by the RoI Align operation.

\subparagraph{Feature comparison}
The projected support feature $\bm{f_{PS}}$ and projected query feature $\bm{f_{PQ}}$ are compared in a point-wise fashion by using $\bm{f_{PS}}$ as a kernel to convolve $\bm{f_{PQ}}$. This convolution has 0 padding added to preserve the spatial information. The output is a score map $\bm{R_0 \in \mathbb{R}^{K \times 1 \times H_Q \times W_Q}}$:

\begin{equation} \label{Feature_comparison}
    \bm{R_0} = \text{conv}(\bm{f_{PQ}}, \text{kernel}), \quad \text{kernel} = \bm{f_{PS}}
\end{equation}

\subparagraph{Score normalisation}
The scores in the score map $\bm{R_0}$ are normalised to prevent extremely large or small values from dominating or destabilising the learning process. It is achieved by Exemplar Normalisation (ENorm), Spatial Normalisation (SNorm), and the element-wise multiplication of their outputs. ENorm normalises $\bm{R_0}$ along the exemplar dimension using $\text{softmax}_{\text{dim}=0}()$, as shown in Equation~\ref{ENorm}, to produce $\bm{R_{EN}}$. Meanwhile, SNorm normalises $\bm{R_0}$ along the height and width dimensions with Equation~\ref{SNorm}, where the $\text{max}_{\text{dim}=(2,3)}()$ finds the maximum value from the corresponding height and width dimensions. The spatially normalised score map $\bm{R_{SN}}$ thus has an important characteristic. The value in the score map at the position most similar or related to the projected support feature $\bm{f_{PS}}$ would be close to 1, whereas the other values range from 0 to 1. The last step of score normalisation is the element-wise multiplication of $\bm{R_{EN}}$ and $\bm{R_{SN}}$ which is presented in Equation~\ref{ENormXSnorm}.

\begin{equation} \label{ENorm}
    \bm{R_{EN}} = \text{softmax}_{\text{dim}=0}(\frac{\bm{R_0}}{\sqrt{H_SW_SC}})
\end{equation}
\begin{equation} \label{SNorm}
    \bm{R_{SN}} = \frac{\text{exp}(\bm{R_0}/\sqrt{H_SW_SC})}{\text{max}_{\text{dim}=(2,3)}(\text{exp}(\bm{R_0}/\sqrt{H_SW_SC}))}
\end{equation}
\begin{equation} \label{ENormXSnorm}
    \bm{R} = \bm{R_{EN}} \otimes \bm{R_{SN}}
\end{equation}
where $\bm{R_0}, \bm{R_{EN}}, \bm{R_{SN}}, \bm{R} \in \mathbb{R}^{K \times 1 \times H_Q \times W_Q}$.

\paragraph{Feature enhancement block}
\label{subsubsec_feature_enhancement}
The aim of feature enhancement block is to exploit the similarity map $\bm{R}$ as weights to enhance the projected query feature $\bm{f_{PQ}}$. This is because the similarity map can effectively represent the relationship between the projected query feature and the projected support feature, but it fails to informatively represent the query image. Feature enhancement block has two steps: weighted feature aggregation and learnable feature fusion.

\begin{figure}[h!]
\centering
\includegraphics[width=.5\linewidth]{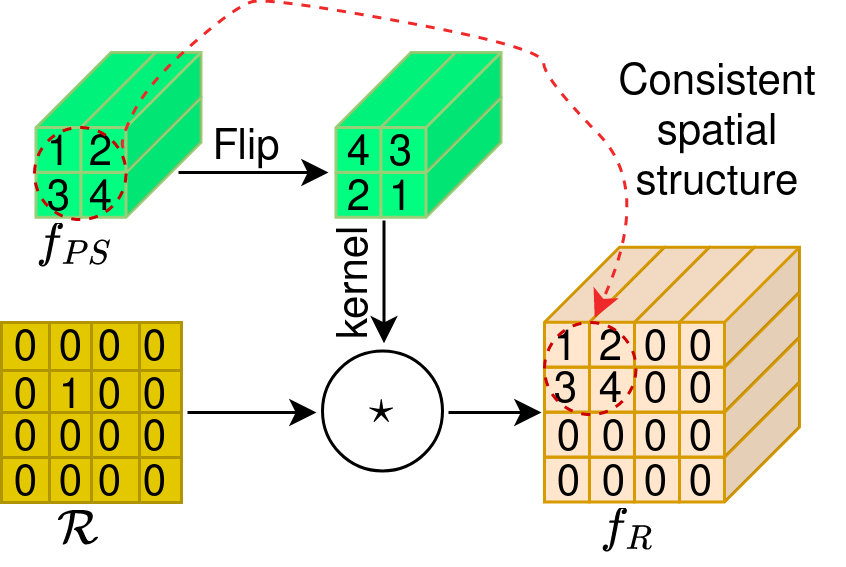}
\caption{\textbf{Illustration of kernel flipping in FEM.} Its purpose is to preserve the spatial structure from the projected support feature $\bm{f_{PS}}$. In this illustration, the $K$ dimension is removed from $\bm{R}$, $\bm{f_{PS}}$, and $\bm{f_R}$ for simplicity, meaning only a support image is used in this example. The motivation behind this design is as follows: suppose that the feature in the projected query feature $\bm{f_{PQ}}$ corresponding to the position of 1 in $\bm{R}$ has the maximum similarity with $\bm{f_{PS}}$, and the other positions in $\bm{f_{PQ}}$ have no similarity. In this case, the similarity-weighted feature $\bm{f_{R}}$ should replicate the values in $\bm{f_{PS}}$ at the position in $\bm{f_{R}}$ corresponding to the position of 1 in $\bm{R}$, whereas other positions in $\bm{f_{R}}$ should be zero.}
\label{fig:kernel_flipping_illustration}
\end{figure}

\subparagraph{Weighted feature aggregation}
Firstly, the projected support feature $\bm{f_{PS}}$ is flipped horizontally and vertically before using it as a kernel to convolve the similarity map $\bm{R}$ with 0 padding added. As illustrated in Figure~\ref{fig:kernel_flipping_illustration}, its purpose is to preserve the spatial structure from the projected support feature $\bm{f_{PS}}$. It is feasible because of the important characteristic gained from the score normalisation step in the feature correlation block. The output is accumulated along the exemplar dimension to produce a similarity-weighted feature $\bm{f_R} \in \mathbb{R}^{C \times H_Q \times W_Q}$ if $K$ support images are used. The weighted feature aggregation is summarised in Equation~\ref{similarity_weighted_feature}. 

\begin{equation} \label{similarity_weighted_feature}
    \bm{f_R} = \text{sum}_{\text{dim}=0}(\text{conv}(\bm{R}, \text{kernel})), \quad \text{kernel} = \text{flip}(\bm{f_{PS}})
\end{equation}

\subparagraph{Learnable feature fusion}
The similarity-weighted feature $\bm{f_R}$ is projected into the same feature space as the projected query feature $\bm{f_{PQ}}$ resides with a $1 \times 1$ convolution whose kernel number is $1$ (since the number of channels of $\bm{f_R}$ is already $C$). This process is followed by a layer normalisation to produce the projected similarity-weighted feature $\bm{f_{PR}} \in \mathbb{R}^{C \times H_Q \times W_Q}$. As shown in Figure~\ref{fig:acfamnetpro_residual_fe_module}, $\bm{f_{PR}}$ is fused into the projected query feature $\bm{f_{PQ}}$ with an efficient network which contains a convolutional block and a layer normalisation to produce the final enhanced feature $\bm{f^{\prime}_Q} \in \mathbb{R}^{C \times H_Q \times W_Q}$. The learnable feature fusion is expressed as:

\begin{equation} \label{Learnable_feature_fusion}
\begin{aligned}
    \bm{f^{\prime}_Q} = \text{LayerNorm}(\bm{f_{PQ}} + \text{h}(\text{LayerNorm}(\text{conv}(\bm{f_R}, \text{kernel})))), \\ 
    \quad \text{kernel} \in \mathbb{R}^{1 \times 1 \times 1}
\end{aligned}
\end{equation}
where the convolutional block $h(x)$ is implemented as:
\begin{equation} \label{Learnable_feature_fusion_conv_block}
\begin{aligned}
    h(x) = \text{conv}(\text{dropout}(\text{LeakyReLU}(\text{conv}(x, \text{kernel}))), \text{kernel}), \\ 
    \quad \text{kernel} \in \mathbb{R}^{C \times 3 \times 3}
\end{aligned}
\end{equation}

\paragraph{Multi-block architecture and comparison with attention}
As illustrated in Figure~\ref{fig:acfamnetpro_residual_fe_module}, the residual feature enhancement module can be stacked $N$ times because the height and width of the enhanced feature map remain the same. This multi-block architecture mimics the multi-head attention mechanism in transformers. Additionally, the vanilla transformer attention mechanism, which is shown in Equation~\ref{Attention_mechanism} where $Q$, $K$, $V$, and $d_k$ are query, key, value, and scale factor, respectively, can be simplified as $\text{similarity}(Q, K)V$. ACFamNet Pro mimics this attention mechanism by interpreting $Q$ as the query feature, $K$ as the support feature, and $V$ as the query feature. Moreover, the feature correlation block and feature enhancement block preserve the spatial information $(C \times H \times W)$. In contrast, the vanilla transformer attention mechanism loses spatial structure because it flattens feature map from $(C \times H \times W)$ to $(C \times HW)$.

\begin{equation} \label{Attention_mechanism}
    \text{Attention}(Q, K, V) = \text{softmax}(\frac{QK^T}{\sqrt{d_k}})V
\end{equation}

\paragraph{Multi-scale support features}
Similar to ACFamNet, the size of support images can be rescaled by different factors, aiming to address the same object of different sizes. This can be achieved by repeating the feature extractor shown in Figure~\ref{fig:acfamnetpro_feature_extractor} with scaled support images and concatenating multiple support features along the first dimension.

\subsubsection{Regression module}
\begin{figure}[h!]
\centering
\includegraphics[width=.8\linewidth]{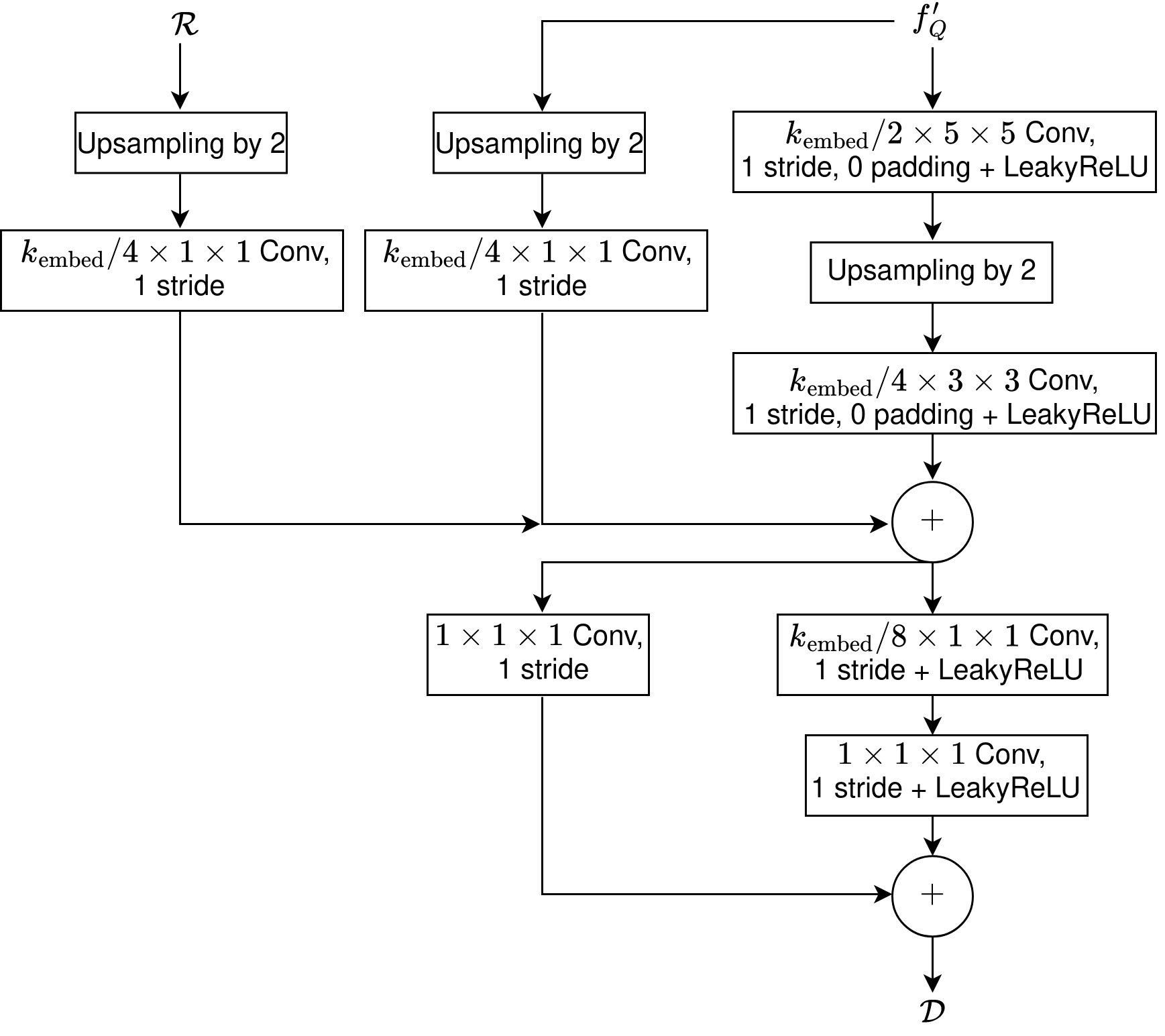}
\caption{Regression module.}
\label{fig:acfamnetpro_regression_module}
\end{figure}
The enhanced feature $\bm{f^{\prime}_Q}$ and similarity map $\bm{R}$ are fed into a regression module to predict the final density map $D \in \mathbb{R}^{H \times W}$, where the height $H$ and width $W$ are identical to the query image. As illustrated in Figure~\ref{fig:acfamnetpro_regression_module}, this regression module consists of four main convolutional layers, where the first convolutional layer is followed by a bilinear upsampling layer to double the spatial resolution. These four convolutional layers are activated by Leaky ReLU activation rather than the traditional ReLU because the former is commonly used for transformer-based models to tackle long sequence data. Additionally, three residual connections are added to the regression module. The enhanced feature $\bm{f^{\prime}_Q}$ and the similarity map $\bm{R}$ are each upsampled, passed through a~1$\times$1~convolution, and added to the input of the third convolutional layer. Finally, the input to the third convolutional layer is added to the output of the final convolutional layer to produce the density map $D$. The number of convolutional kernels and the kernel size in each convolutional layer are detailed in Figure~\ref{fig:acfamnetpro_regression_module}, where $k_{\text{embed}}$ is a hyper-parameter.

\subsubsection{Comparison with SAFECount}
The design of ACFamNet Pro is inspired by SAFECount with three distinct differences. Firstly, ACFamNet Pro uses a simple learnable convolutional layer to extract features with RoI Align, whereas SAFECount uses the first three frozen blocks of ResNet-18 to extract features with RoI pooling. Secondly, although the feature enhancement module in ACFamNet Pro is largely similar to that of SAFECount, it additionally incorporates the similarity map $\bm{R}$ as a residual connection to enhance density map estimation. Finally, the regression module in ACFamNet Pro is completely redesigned with three residual connections to further improve density map estimation.

\section{Experiments}
\label{sec-exp}
This section first introduces the training and evaluation strategies as they are the same for both ACFamNet and ACFamNet Pro. It then presents a series of experiments to separately assess if each model achieves the aim of this work. These experiments include hyper-parameter tuning, ablation studies, comparisons with other counting methods, and cross-category adaptation.

\subsection{Training and evaluation strategies}
\label{subsec-train-eva-strategies}
\subsubsection{Evaluation metrics}
Mean Absolute Error (MAE) and Root Mean Square Error (RMSE) are widely used metrics for counting tasks~\citep{wan_adaptive_2019, ranjan_learning_2021}. They are defined as $\text{MAE} = \frac{1}{N}\sum_{i=1}^{N}|\hat{y}_{i} - y_{i}|$ and $\text{RMSE} = \sqrt{\frac{1}{N}\sum_{i=1}^{N}(\hat{y}_{i} - y_{i})^{2}}$, where $N$, $\hat{y_i}$, and $y_i$ denote the number of samples, the predicted count, and the ground truth count, respectively. However, plate images containing a large number of colonies can generate disproportionately large errors, which may dominate the MAE and RMSE. This limitation may therefore bias the evaluation and prevent these metrics from accurately reflecting overall counting performance. To overcome this issue, the absolute error is normalised by the ground truth count to generate a normalised absolute error for each sample. The normalised absolute errors for all predictions are averaged to produce the Mean Normalised Absolute Error (MNAE), which is less sensitive to variations in colony count. It is defined as $\text{MNAE} = \frac{1}{N}\sum_{i=1}^{N} |\frac{\hat{y}_{i} - y_{i}}{y_{i}}|$.

\subsubsection{Data}
Experiments are conducted on the Synoptics Dataset introduced in \S~\ref{subsec-data}. The training set, which is 80\% of the whole dataset, is used for model development and evaluation through $k$-fold cross-validation. After identifying the best-performing hyper-parameters, the model is retrained from scratch on the entire 80\% training set. Finally, the remaining 20\% test set, which is reserved as a held-out test set, is used to assess the final model's performance.

Following the setup introduced by~\citet{ranjan_learning_2021}, the ground truth density map for the Synoptics Dataset is generated using a Gaussian function. The Gaussian mean is defined as the average distance between each dot and its nearest neighbour for the whole map. The Gaussian standard deviation is set to a quarter of the kernel size. This specific design adapts the ground truth function to each individual image. As a result, the density map is tailored to variations in image size and object density.

\subsubsection{Training strategy}
\label{subsec_training_strategy}
$K$-fold cross-validation is used to train and estimate the performance of different neural networks with different hyper-parameters. In this procedure, the training data are equally divided into $k$ folds, and the model is trained and validated $k$ times. At each time, different $k-1$ folds and the remaining fold are used for training and validation, respectively. The final performance is obtained by averaging the results across all $k$ validation folds. According to~\citet{hastie_elements_2009}, 5 and 10 are common choices for $k$. However, it is set to 5 in this work because the Synoptics training set contains 100 samples, and 5 is a suitable divisor that ensures balanced splits while avoiding excessive computational cost. Meanwhile, choosing a larger divisor of the training set size, such as $k=10$, will significantly increase training time.

To further conserve computational resources, this work focuses on hyper-parameters that determine the network architecture rather than those related to the training process. Another reason for this choice is that training-related hyper-parameters can often adopt well-established values from prior studies. In contrast, architectural hyper-parameters are more closely tied to the specific design of the proposed network, and therefore require dedicated investigation.

Before training, images in the 4 training folds are normalised by subtracting the mean of pixel values of those folds and dividing by their standard deviation to speed up convergence for training. The mean and standard deviation of pixel values in the 4 training folds are reserved to normalise images in the corresponding validation fold during evaluation. This data normalisation procedure has been widely used in neural network training~\citep{Krizhevsky2012, He2015}.

Mean Squared Error (MSE) is adopted as the loss function since models in this work are formulated as regression tasks. The models are trained using the Adam optimiser~\citep{kingma_adam_2014} with a learning rate of $10^{-5}$, a batch size of 1, and 1500 epochs, which are identical to the choices made by~\citet{ranjan_learning_2021}. Additionally, early stopping is applied if the validation MNAE does not improve by at least 1\% for 200 consecutive epochs. This strategy avoids unnecessary computation when performance plateaus.

\subsection{Experiments on ACFamNet}
\label{subsec-acfamnet-exp}
\subsubsection{Training}
Experiments in this section follow the same setup explained in \S~\ref{subsec-train-eva-strategies} for data, training, and evaluation. Before training, parts of neural networks are initialised with parameters from previously trained models. Specifically, the first convolutional layer of ACFamNet shown in Figure~\ref{fig:acfamnet_feature_correlation_module} is initialised with the corresponding layer from ResNet-50 to leverage its learned feature representations. Similarly, the density map prediction module in ACFamNet is pre-trained on the FSC-147 dataset~\citep{ranjan_learning_2021}, which consists of 6135 images across 147 object categories, enabling it to transfer general counting knowledge. During hyper-parameter tuning, if the first convolutional layer has more than 64 kernels, the pre-trained kernels are duplicated to match the required dimensionality. For models with a scale factor is 1, only a subset of the pre-trained parameters is transferred to ensure dimensional compatibility.

\subsubsection{Hyper-parameter tuning}
\label{subsec_para_tuning_acfamnet}
\paragraph{Setup}
Hyper-parameter tuning is performed on ACFamNet to find the optimal configuration for maximising counting performance. These hyper-parameters include the number of kernels in the first convolutional kernel, the RoI Align output size, and the number of scales. Among them, the number of kernels in the first convolutional layer, i.e. the value of $k$ in Figure~\ref{fig:acfamnet_feature_correlation_module}, is fine-tuned with 64, 128, 256, and 512, as they are commonly used in the design of convolutional kernels. The RoI Align output size is fine-tuned with~1$\times$1,~3$\times$3,~5$\times$5, and~7$\times$7~because an odd RoI Align output size can provide a symmetrical kernel for feature correlation. Additionally, the number of scales is fine-tuned with 1 and 3. The 1 scale factor indicates no scaling (i.e. Blocks 2 and 3 are removed in Figure~\ref{fig:acfamnet_feature_correlation_module}), whereas the 3 scale factors consist of 1, 0.9, and 1.1 (i.e. $s_1=0.9$ and $s_2=1.1$ in Figure~\ref{fig:acfamnet_feature_correlation_module}). The selection of 3 scale factors is identical to FamNet.

\paragraph{Results}
Table~\ref{acfamnet_para_tuning_results} presents the results of tuning ACFamNet, with each validation MNAE reported as the mean and standard deviation obtained from $5$-fold cross-validation. The results show that ACFamNet tends to achieve slightly lower validation MNAE with 1 scale factor than with 3 scale factors, regardless of the RoI Align output size or the number of convolutional kernels used. This tendency may be due to the nature of small objects, where resizing them with different scale factors increases the feature space that is not useful, ultimately degrading counting performance.

\begin{table}[h!]
\centering
\begin{threeparttable}
\caption{Hyper-parameter tuning results for ACFamNet.}
\label{acfamnet_para_tuning_results}
\begin{tabular}{lccc}
        \hline
                                                           & \multirow{2}{*}{RoI Align} & \multicolumn{2}{c}{Validation MNAE(\%)}                         \\ \cline{3-4}
                                                           &                            & 3 Scale Factors                         & 1 Scale Factor        \\ \hline
        \multirow{4}{*}{\rotatebox[origin=c]{90}{k = 64}}  & 1$\times$1               & 99.99 $\pm$ 0.10\tnote{a}               & 99.98 $\pm$ 0.03      \\
                                                           & 3$\times$3               & 13.29 $\pm$ 2.33                        & 12.06 $\pm$ 2.48      \\
                                                           & 5$\times$5               & 14.06 $\pm$ 1.75                        & 12.86 $\pm$ 2.39      \\
                                                           & 7$\times$7               & 17.21 $\pm$ 1.88                        & 15.63 $\pm$ 1.63      \\ \cline{1-4}
        \multirow{4}{*}{\rotatebox[origin=c]{90}{k = 128}} & 1$\times$1               & 99.98 $\pm$ 0.02                        & 31.49 $\pm$ 34.37     \\
                                                           & 3$\times$3               & 12.23 $\pm$ 2.15                        & 12.28 $\pm$ 1.85      \\
                                                           & 5$\times$5               & 14.60 $\pm$ 2.45                        & 13.69 $\pm$ 2.28      \\
                                                           & 7$\times$7               & 17.24 $\pm$ 2.40                        & 15.30 $\pm$ 1.35      \\ \cline{1-4}
        \multirow{4}{*}{\rotatebox[origin=c]{90}{k = 256}} & 1$\times$1               & 14.27 $\pm$ 3.93                        & 13.52 $\pm$ 1.30      \\
                                                           & 3$\times$3               & 12.94 $\pm$ 1.91                        & \bf{11.85 $\pm$ 2.53} \\
                                                           & 5$\times$5               & 14.13 $\pm$ 2.74                        & 13.91 $\pm$ 2.01      \\
                                                           & 7$\times$7               & 16.95 $\pm$ 2.45                        & 16.18 $\pm$ 1.19      \\ \cline{1-4}
        \multirow{4}{*}{\rotatebox[origin=c]{90}{k = 512}} & 1$\times$1               & 13.46 $\pm$ 3.16                        & 13.50 $\pm$ 1.96      \\
                                                           & 3$\times$3               & 13.78 $\pm$ 2.12                        & 13.60 $\pm$ 2.69      \\
                                                           & 5$\times$5               & 15.93 $\pm$ 3.08                        & 14.74 $\pm$ 2.05      \\
                                                           & 7$\times$7               & 17.64 $\pm$ 1.57                        & 16.22 $\pm$ 3.00      \\ \hline
\end{tabular}
\begin{tablenotes}
        \item[a] Each validation MNAE presented in this table includes both the mean and standard deviation computed from $5$-fold cross-validation. A lower mean MNAE indicates better performance.
\end{tablenotes}
\end{threeparttable}
\end{table}

The increase of RoI Align output size from 3$\times$3~to~7$\times$7 consistently leads to higher validation MNAE, regardless of the number of scale factors or convolutional kernels used. However, this trend does not hold for the~1$\times$1~RoI Align output size. When the RoI Align output size is~1$\times$1~and the kernel number is 64 or 128, the validation MNAE is almost close to 100\%, indicating extremely poor performance. Interestingly, the validation MNAE drops substantially when the number of kernels exceeds 256, even though the RoI Align output size is~1$\times$1. One possible explanation is that increasing the number of kernels enhances ACFamNet's capacity to capture more complex features, compensating for the limited spatial information of the~1$\times$1~RoI Align output size. Increasing the number of kernels in the first convolutional layer greatly raises computational cost but yields only minor fluctuations in validation MNAE, regardless of the RoI Align output size or scale factors used. These results suggest that the performance gains are trivial, even if they come at the cost of increasing the number of kernels.

The best result from the hyper-parameter tuning is obtained with 256 kernels, a~3$\times$3~RoI Align output size, and 1 scale factor, producing a mean validation MNAE of 11.85\% with a standard deviation of 2.53\%. This result indicates that ACFamNet, which is modified from FamNet, can effectively count small and clustered colonies because 11.85\% is a relatively low error rate. Notably, only three labelled exemplars are randomly selected for training, demonstrating that ACFamNet can successfully learn from limited labelled data. The detailed $5$-fold cross-validation results are provided in Appendix Table~\ref{table_detailed_5_cs_results_from_best_acfamnet}. Overall, the fine-tuned ACFamNet generalises well to unseen data because most of the validation metrics are lower than the corresponding training metrics. Particularly, the training MNAE is higher than the validation MNAE. One possible explanation is that a smaller dataset has lower intrinsic variance, meaning ACFamNet captures the complexity of the data and the intrinsic variance of the training set is higher than that of the validation set. 

The prediction results for two example validation images are illustrated in Appendix Figures~\ref{fig:ACFamNet_image_66_cropped} and~\ref{fig:ACFamNet_image_122_cropped}. Both figures show that ACFamNet's predictions (91.12 vs 94 and 139.94 vs 142) are close to the ground truth counts, demonstrating that the model can effectively count small and clustered colonies. Enlarged illustrations of all predicted density maps without overlays are available in~\citep{zheng_learning_2024}.

\subsubsection{Ablation studies}
\paragraph{Setup}
Two ablation studies are conducted with the Synoptics Dataset to analyse the effectiveness of different ACFamNet components and the impact of the number of exemplars on counting performance. ACFamNet used in the first ablation study is the one with three exemplars and optimised hyper-parameters (256 kernels). The components are the single scale factor and RoI Align because they are the distinctive design elements of ACFamNet. When ACFamNet includes the RoI Align component, its RoI Align output size is set to~3$\times$3, and otherwise it is replaced with RoI pooling. Similarly, when the single scale factor component is excluded, it is replaced with 3 scale factors which are $1$, $0.9$, and $1.1$.\footnote{ACFamNet without the single scale factor component is equivalent to the architecture illustrated in Figure~\ref{fig:acfamnet_feature_correlation_module}. In contrast, ACFamNet with the single scale factor component is equivalent to removing Blocks 2 and 3 in Figure~\ref{fig:acfamnet_feature_correlation_module}.} In the second ablation study, ACFamNet is also the one with optimised hyper-parameters (256 kernels,~3$\times$3 RoI Align output size, and $1$ scale factor). The data, training method, and evaluation method used in both studies are identical to those introduced in \S~\ref{subsec-train-eva-strategies}.

\paragraph{Results}
\begin{table*}[t]
        \centering
        \caption{Analysis of the effectiveness of different ACFamNet components.}
        \label{tab_acfamnet_ablation_study}
    \begin{tabular}{>{\centering}p{0.3\textwidth}>{\centering}p{0.1\textwidth}>{\centering}p{0.1\textwidth}>{\centering}p{0.1\textwidth}>{\centering\arraybackslash}p{0.1\textwidth}}
                \hline
                Component          & \multicolumn{4}{c}{Combination}                                                    \\ \hline
                Single scale factor & \texttimes                       & \texttimes     & \checkmark     & \checkmark     \\
                RoI Align           & \texttimes                       & \checkmark     & \texttimes     & \checkmark     \\ \hline
                Validation MNAE(\%) & 17.73 $\pm$ 3.37                   & 12.94 $\pm$ 1.91 & 20.39 $\pm$ 4.43 & \bf{11.85 $\pm$ 2.53} \\ \hline
        \end{tabular}
\end{table*}

The first ablation study results in Table~\ref{tab_acfamnet_ablation_study} highlight the importance of ACFamNet's individual components, namely the single scale factor and RoI Align. The combination of these two components further reduces the validation MNAE from 17.73\% to 11.85\%. Table~\ref{tab_acfamnet_num_exemplar_results} shows that ACFamNet's counting performance improves with the increase of exemplars. ACFamNet can even produce reasonable counting results with only a single exemplar. These two patterns are similar to those discovered by FamNet's authors, suggesting ACFamNet is an effective adaptation of FamNet for counting small and clustered colonies.

\begin{table}[h!]
        \centering
        \caption{Performance of ACFamNet trained with different numbers of exemplars.}
        \label{tab_acfamnet_num_exemplar_results}
    \begin{tabular}{>{\centering}p{0.15\textwidth}>{\centering\arraybackslash}p{0.3\textwidth}}
                \hline

                                                                   Exemplar         & Validation MNAE (\%)                              \\ \hline
                                                                   1                           & 14.94 $\pm$ 2.32                               \\
                                                                   2                           & 13.07 $\pm$ 2.35                              \\
                                                                   3                           & \bf{11.85 $\pm$ 2.53}                              \\ \hline
        \end{tabular}
\end{table}

\subsubsection{Comparison with FamNet}
\paragraph{Setup}
It is necessary to compare ACFamNet with FamNet because the former is inspired by the latter. The vanilla FamNet has 3 scale factors, RoI pooling, and a complex non-trainable feature extraction module. FamNet is examined with 3 scale factors and 1 scale factor. Additionally, FamNet is fine-tuned with different RoI Align operations. The data, training method, and evaluation method are identical to those introduced in \S~\ref{subsec-train-eva-strategies}.

\paragraph{Results}
The vanilla FamNet can count small bacterial colonies with a mean MNAE of 22.33\% and a standard deviation of 6.53\% based on $5$-fold cross-validation. Reducing the number of scale factors for FamNet from 3 to 1 slightly degrades performance, resulting in a mean MNAE of 23.83\% with a standard deviation of 6.66\%, as shown in Appendix Table~\ref{tab_famnet_with_diff_scale_factors}. This decline may be attributed to its non-trainable feature extraction module, which may require additional features to compensate. Notably, both two standard deviations are higher than those in Table~\ref{acfamnet_para_tuning_results}, suggesting FamNet is less stable than ACFamNet. Detailed $5$-fold cross-validation results of the vanilla FamNet in Appendix Table~\ref{table_detailed_5_cs_results_from_best_famnet} further reveal that FamNet's ability to generalise to unseen data is less stable. This is observed from the pattern that the validation MNAE has a higher standard deviation than the corresponding training MNAE across $5$-fold datasets.

\begin{table}[h!]
\centering
\caption{Results of tuning RoI Align output size for FamNet.}
\label{tab_famnet_with_diff_roi_align}
\begin{tabular}{>{\centering}p{0.17\textwidth}>{\centering}p{0.2\textwidth}>{\centering\arraybackslash}p{0.2\textwidth}}
    \hline
    \multirow{2}{*}{RoI Align}&\multicolumn{2}{c}{Validation MNAE (\%)} \\\cline{2-3}
               & 3 Scale Factors              & 1 Scale Factor           \\ \hline
    1$\times$1 & 100.0 $\pm$ 0.00      & 100.0 $\pm$ 0.00  \\
    3$\times$3 & \bf{25.29 $\pm$ 5.60} & 46.14 $\pm$ 27.31 \\
    5$\times$5 & 84.38 $\pm$ 30.38     & 46.76 $\pm$ 26.90 \\
    7$\times$7 & 99.92 $\pm$ 0.03      & 68.14 $\pm$ 25.83 \\ \hline
\end{tabular}
\end{table}

\begin{table}[h!]
\centering
\caption{Comparison between ACFamNet and the vanilla FamNet.}
\label{tab_acfamnet_vs_famnet}
\begin{tabular}{cc}
        \hline

                                                           Model         & Validation MNAE (\%)                                            \\ \hline
                                                           ACFamNet                         & \bf{11.85} $\pm$ \bf{2.53}                             \\
                                                           FamNet                           & 22.33 $\pm$ 6.53                              \\ \hline
\end{tabular}
\end{table}

Changing RoI Align output size has yet to improve FamNet's counting performance because all MNAE values in Table~\ref{tab_famnet_with_diff_roi_align} exceed the vanilla FamNet's 22.33\% MNAE. This could be attributed to the non-learnable feature extraction module, suggesting that FamNet needs to learn from data to effectively handle the interpolated features from RoI Align operations. Finally, the fine-tuned ACFamNet outperforms the vanilla FamNet by 10.48\% in validation MNAE as presented in Table~\ref{tab_acfamnet_vs_famnet}.

\subsubsection{Comparison with traditional methods}
\label{subsec_compa_acfamnet_vs_traditional_methods}
\paragraph{Setup}
In order to compare machine learning-based counting methods with traditional counting methods, OpenCFU~\citep{Geissmann2013} and AutoCellSeg \citep{Khan2018} are selected alongside ACFamNet. OpenCFU and AutoCellSeg are popular open-source tools that are based on traditional image thresholding algorithms which require users to pre-define some parameters for object detection. These three methods are evaluated on the same Synoptics test set that has never been used to ensure a fair comparison. In other words, the optimised ACFamNet (256 kernels,~3$\times$3~RoI Align output size, and 1 scale factor) is trained on the Synoptics training set and evaluated on its test set using a hold-out evaluation. The training of ACFamNet follows the same learning rate, batch size, epoch number, early stopping strategy, loss function, and Adam optimiser described in~\S\ref{subsec_training_strategy}. The comparison of running time for these methods is not included in this work because deep learning-based methods outperform traditional methods due to shared computation for batched inputs.

\paragraph{Results}
\begin{table}[h!]
\centering
\caption{Comparison between ACFamNet and traditional counting methods.}
\label{tab_acfamnet_vs_opencfu_autocellseg}
\begin{tabular}{cccc}
    \hline
    Metric                        & OpenCFU & AutoCellSeg & ACFamNet                                            \\ \hline
    MAE                           & 41.12        & 60.92        & 11.54                             \\
    RMSE                          & 47.76        & 69.87        & 15.56                             \\
    MNAE (\%)                     & 46.57        & 68.73        & \bf{12.52}                              \\ \hline
\end{tabular}
\end{table}

As shown in Table~\ref{tab_acfamnet_vs_opencfu_autocellseg}, ACFamNet surpasses traditional counting methods, achieving 12.52\% MNAE, which is 34.05\% and 56.21\% lower than OpenCFU and AutoCellseg, respectively. Detailed hold-out evaluation results in Appendix Table~\ref{tab_acfamnet_final_hold_out_result} further show that ACFamNet generalises well to unseen data. Similar to the previous $5$-fold cross-validation results in Appendix Table~\ref{table_detailed_5_cs_results_from_best_acfamnet}, performance on unseen data is better than that on the training set. This might be due to the same reason that ACFamNet captures the complexity of the data and the intrinsic variance of the training set is higher than that of the test set, as detailed in Table~\ref{tab_data_v2_mean_std_var}.

\begin{figure}[h!]
\centering
\includegraphics[width=.45\linewidth]{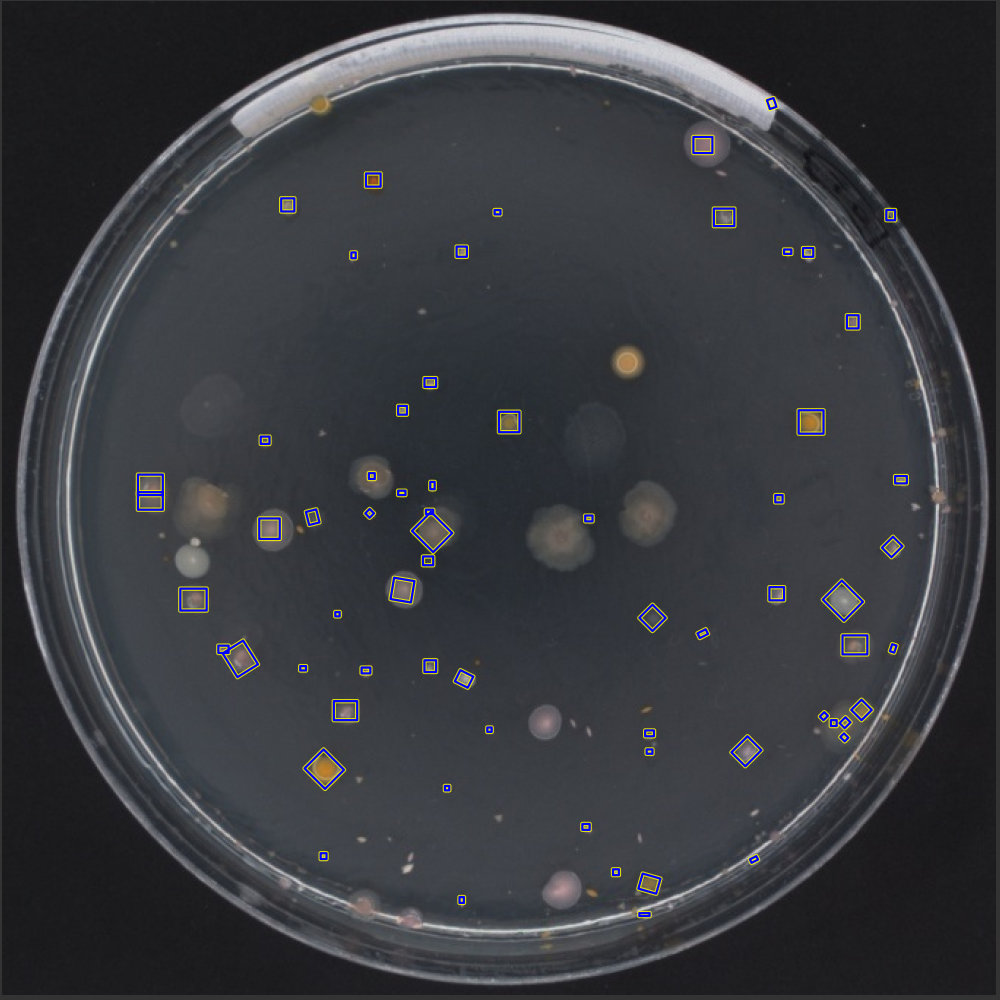}
\includegraphics[width=.45\linewidth]{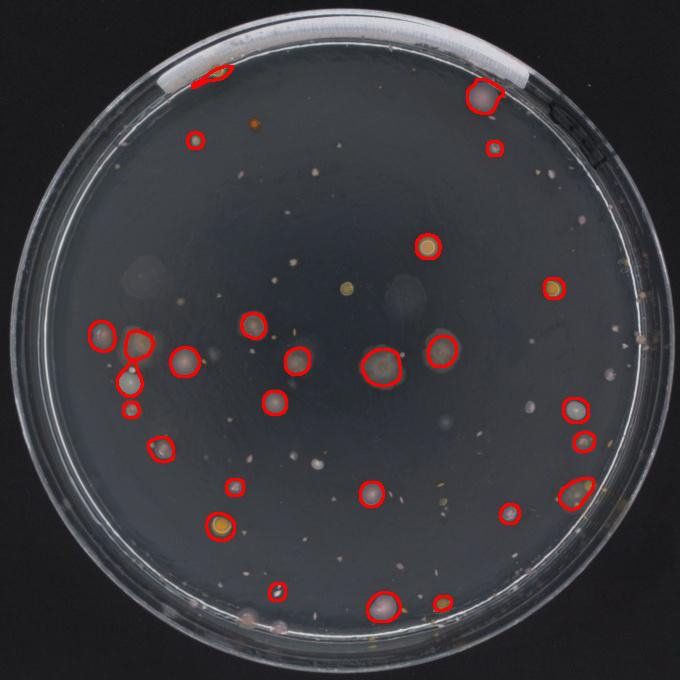}
\caption{Counting results for an image with 83 colonies. Left: OpenCFU detects 66 colonies. Right: AutoCellSeg detects 27 colonies.}
\label{fig:63_from_traditional_methods}
\end{figure}

\begin{figure}[h!]
\centering
\includegraphics[width=.90\linewidth]{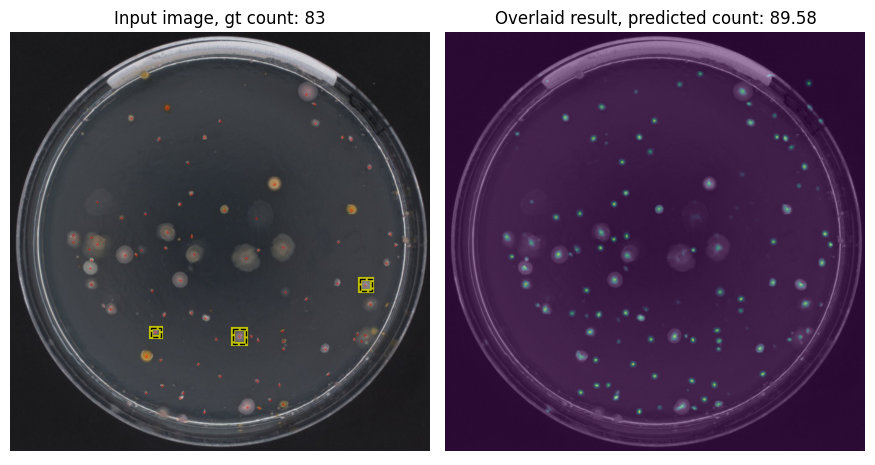}
\caption{Counting results for an image with 83 colonies. ACFamNet detects 89.58 colonies.}
\label{fig:63_from_ACFamNet}
\end{figure}

An analysis of prediction results from these three methods is conducted. Figure~\ref{fig:63_from_traditional_methods} shows that traditional counting methods fail to detect colonies with varying colour, size, shape, and density. One possible explanation is the nature of traditional image processing algorithms, such as thresholding, which rely heavily on manual parameter adjustment to handle high variability. Contrary to traditional counting methods, ACFamNet, which is based on machine learning, can effectively learn from data to tackle this variability as illustrated in Figure~\ref{fig:63_from_ACFamNet}. Moreover, ACFamNet's neural network shares computation when processing large volumes of data, providing high scalability for use in real laboratory settings.

\subsubsection{Category adaptation}
\label{subsec_acfamnet_cross_category_adaptation}
\paragraph{Setup}
This experiment aims to evaluate how well ACFamNet can count small and clustered colonies from a different category. FamNet, which is based on few-shot learning, can count objects of a different category as long as three exemplars are provided. It is therefore naturally hypothesised that ACFamNet, a modified version of FamNet, should also be capable of counting objects from a different category.

\begin{figure}[h!]
\centering
\includegraphics[width=.20\linewidth]{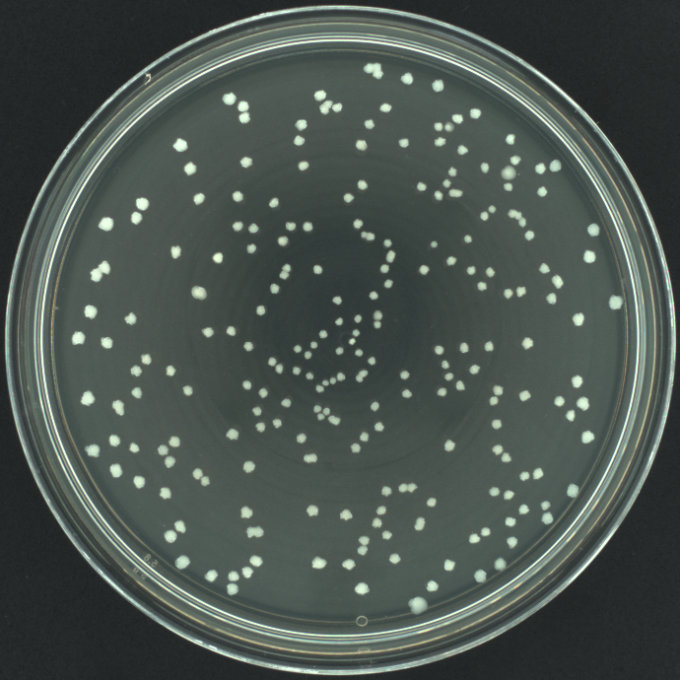}
\includegraphics[width=.20\linewidth]{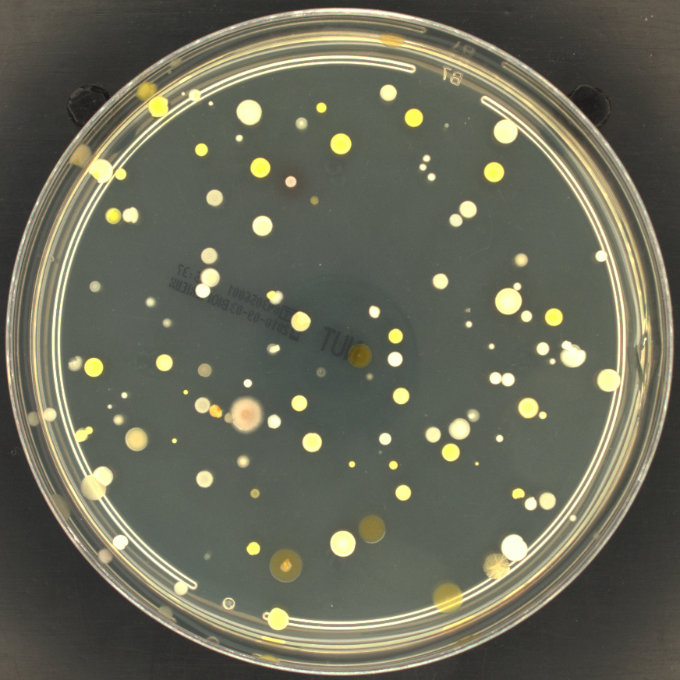}
\includegraphics[width=.20\linewidth]{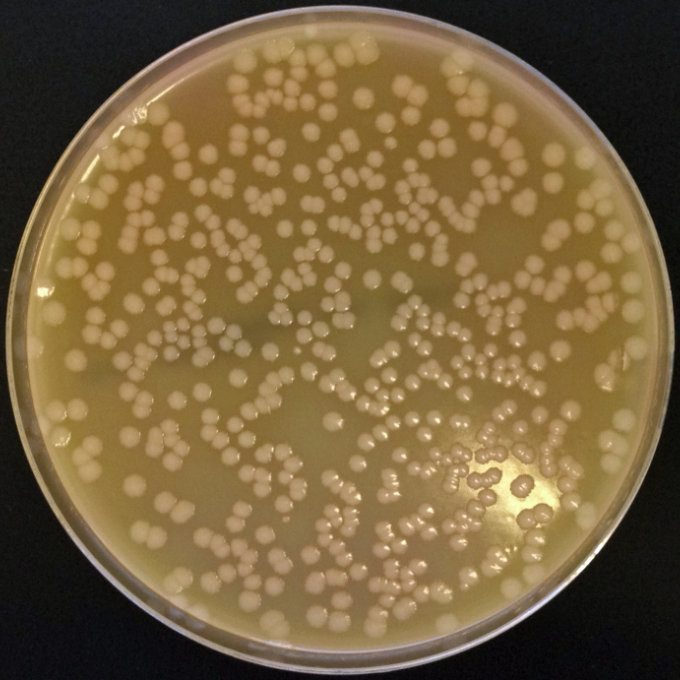}
\includegraphics[width=.20\linewidth]{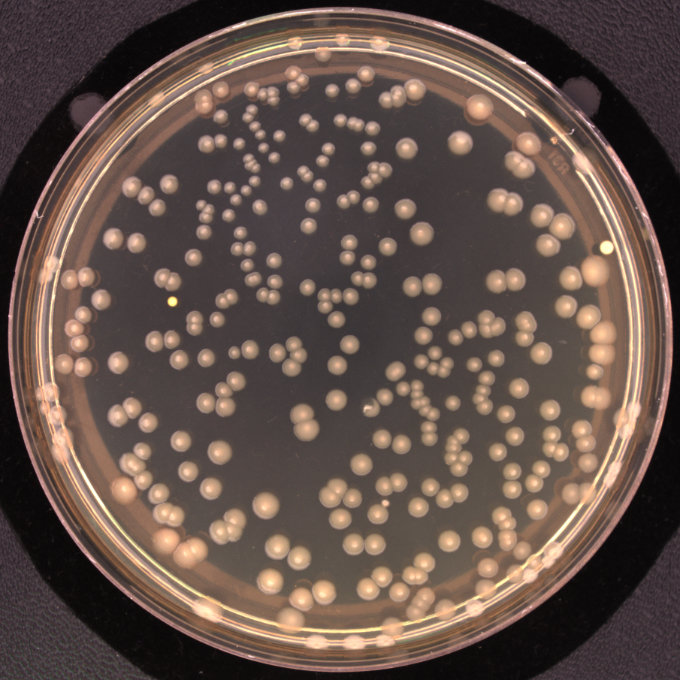}
\caption{Four plate images with colonies that are completely different from the Synoptics Dataset. Left to right: Plate Image A, B, C, and D. Enlarged illustrations are presented in Figures~\ref{fig:2.8_prediction_acfamnet},~\ref{fig:envi_image_prediction_acfamnet},~\ref{fig:IMG_0431_prediction_acfamnet}, and~\ref{fig:pour04_prediction_acfamnet}.}
\label{fig:new_colony_species_two_images}
\end{figure}

Four plate images containing colony species different from those in the training data are used, as shown in Figure~\ref{fig:new_colony_species_two_images}. The differences between these images and the Synoptics Dataset images in Figure~\ref{example-plate-img1} are significant in terms of colour, background, density, and species. The ACFamNet used in this experiment is the one obtained from \S~\ref{subsec_compa_acfamnet_vs_traditional_methods}, i.e. the ACFamNet retrained on the whole Synoptics training set with optimised hyper-parameters. This approach avoids repetitive model training while taking full advantage of all available training data.

\paragraph{Results}
\begin{table}[h!]
\centering
\caption{Results of ACFamNet's cross-category prediction.}
\label{tab_acfamnet_cross_category_prediction}
\begin{tabular}{ccc}
        \hline
        Plate image        & Ground truth & ACFamNet \\ \hline
        A   & 228                & 306.31 \\
        B   & 124                & 475.85 \\
        C & 529                & 3.1  \\
        D  & 302                & 832.12                   \\ \hline  \hline
        MAE               & RMSE               & MNAE (\%)                \\ \hline
        371.55            & 414.58             & 148.26 \\ \hline
\end{tabular}
\end{table}

Table~\ref{tab_acfamnet_cross_category_prediction} presents detailed results of ACFamNet's cross-category prediction. The MNAE from the four plage images is 148.26\%, indicating ACFamNet cannot readily generalise to colonies of a different category. One possible reason is that ACFamNet is trained on the Synoptics Dataset in which each plate image may contain colonies of the same species, violating a key assumption of few-shot learning. It is also likely caused by the limited number of images and the significant change of colour in the plate area. Despite the high error, the detailed predictions illustrated in Appendix Figures~\ref{fig:2.8_prediction_acfamnet},~\ref{fig:envi_image_prediction_acfamnet},~\ref{fig:IMG_0431_prediction_acfamnet}, and~\ref{fig:pour04_prediction_acfamnet} imply that ACFamNet can still predict the location of different colonies based on the three exemplars. This is because the location of predicted dots in the predicted density map has a tendency to match the location of colonies in the input image.

\subsubsection{Summary}
This section has introduced ACFamNet, which is a special adaptation of FamNet, to count small and clustered colonies. The fine-tuned ACFamNet achieves a mean validation MNAE of 11.85\% with a standard deviation of 2.53\% through $5$-fold cross-validation on the Synoptics training set. Ablation studies reveal that the single scale factor and RoI Align are ACFamNet's important components, as removing them degrades performance to a mean validation MNAE of 17.73\% under the same $5$-fold cross-validation setup. With a mean validation MNAE of 22.33\%, FamNet performs worse than ACFamNet by 10.48\%, demonstrating ACFamNet's superior accuracy. FamNet is also evaluated with different scale factors and RoI Align operations with a finding that these two components are detrimental, possibly due to its overly complicated and non-learnable feature extraction module.

ACFamNet is further evaluated on the Synoptics test set after being trained from scratch on the Synoptics training set using a hold-out evaluation. The test MNAE is 12.52\%, representing a significant improvement over traditional counting methods such as OpenCFU and AutoCellSeg, which generate 46.57\% and 68.73\% in MNAE, respectively. Finally, ACFamNet is evaluated on four plate images containing colonies of completely different species, yielding a high MNAE of 148.26\%. This excessive error rate may be attributed to the dataset on which ACFamNet is trained. Contributing factors include the overlap of colony categories within training images, the limited number of training images, and the significant colour variation in plate areas between training images and these four images.

\subsection{Experiments on ACFamNet Pro}
\label{subsec-acfamnetpro-exp}
\subsubsection{Training}
\label{subsec_acfamnet_pro_training}
ACFamNet Pro is trained on the Synoptics training set through the same $5$-fold cross-validation as ACFamNet. The evaluation metrics, MSE loss function, Adam optimiser, learning rate, batch size, epoch number, and early stopping strategy remain unchanged. Additionally, the weights of the feature extractor illustrated in Figure~\ref{fig:acfamnetpro_feature_extractor} are initialised using a zero-mean Gaussian distribution with a standard deviation of $0.01$, following the same strategy as SAFECount.

\subsubsection{Hyper-parameter tuning}
\label{subsec_para_tuning_acfamnetpro}
\paragraph{Setup}
Hyper-parameter tuning is conducted on ACFamNet Pro, with a focus on the RoI operation and the scale factors related to the backbone feature extractor. Specifically, the backbone with 128 kernels, which is the same number of kernels used in ACFamNet, is evaluated with both frozen and unfrozen (learnable) weights. The RoI operation with a~3$\times$3~output size, which is the optimal size identified in ACFamNet, is fine-tuned with RoI Align and RoI pooling. Similar to ACFamNet, the number of scales is fine-tuned with 1 and 3. The projected feature dimension is 256, the number of residual feature enhancement modules is 4, and the embed dimension $k_{embed}$ in the regression module is set to 1024. Because these hyper-parameters are optimal in SAFECount, they are not tuned in ACFamNet Pro.

\paragraph{Results}
\begin{table*}[t]
	\centering
	\caption{Hyper-parameter tuning results for ACFamNet Pro.}
	\label{acfamnet_pro_para_tuning_results}
	\begin{tabular}{ccccc}
		\hline
		                                                     &                                                     & \multirow{2}{*}{RoI (3$\times$3)} & \multicolumn{2}{c}{Validation MNAE(\%)}                                     \\ \cline{4-5}
		                                                     &                                                     &                                   & 3 Scale Factors                         & 1 Scale Factor                    \\ \hline
		\multirow{4}{*}{\rotatebox[origin=c]{90}{Learnable}} & \multirow{4}{*}{\rotatebox[origin=c]{90}{backbone}} & \multirow{2}{*}{RoI pooling}      & \multirow{2}{*}{10.76 $\pm$ 3.35}       & \multirow{2}{*}{11.23 $\pm$ 2.94} \\
		                                                     &                                                     &                                   &                                         &                                   \\ \cline{3-5}
		                                                     &                                                     & \multirow{2}{*}{RoI Align}        & \multirow{2}{*}{\bf{9.62 $\pm$ 3.35}}   & \multirow{2}{*}{10.27 $\pm$ 3.72} \\
		                                                     &                                                     &                                   &                                         &                                   \\ \hline
		\multirow{4}{*}{\rotatebox[origin=c]{90}{Frozen}}    & \multirow{4}{*}{\rotatebox[origin=c]{90}{backbone}} & \multirow{2}{*}{RoI pooling}      & \multirow{2}{*}{12.18 $\pm$ 3.12}       & \multirow{2}{*}{11.61 $\pm$ 3.55} \\
		                                                     &                                                     &                                   &                                         &                                   \\ \cline{3-5}
		                                                     &                                                     & \multirow{2}{*}{RoI Align}        & \multirow{2}{*}{10.85 $\pm$ 1.86}       & \multirow{2}{*}{11.52 $\pm$ 2.82} \\
		                                                     &                                                     &                                   &                                         &                                   \\ \hline
	\end{tabular}
\end{table*}

Table~\ref{acfamnet_pro_para_tuning_results} presents the hyper-parameter tuning results for ACFamNet Pro. The model consistently performs better with a learnable backbone than with a frozen one, regardless of the choice of RoI operation or number of scale factors. This can be attributed to the same factor that enables the end-to-end trainable ACFamNet to outperform FamNet, namely, that the model's modules become differentiable and easier to optimise for the entire task. In addition, ACFamNet Pro tends to perform better when RoI Align is used, regardless of the backbone configuration or scale factors, likely because RoI Align mitigates feature misalignment. Contrary to ACFamNet, ACFamNet Pro performs better when the 3 scale factors are used. One possible explanation is that the multi-head attention mechanism in ACFamNet Pro requires larger feature space which is provided by additional scale factors.

The best hyper-parameter tuning result is achieved by using a learnable backbone, RoI Align, and 3 scale factors, producing a mean validation MNAE of 9.62\% with a standard deviation of 3.35\%. Compared with ACFamNet's best performance (11.85\% MNAE) in Table~\ref{acfamnet_para_tuning_results}, ACFamNet Pro outperforms ACFamNet by 2.23\% in the mean validation MNAE. The detailed $5$-fold cross-validation results in Appendix Table~\ref{table_detailed_5_cs_results_from_best_acfamnetpro} show that the fine-tuned ACFamNet Pro can generalise well to unseen data. Similar to ACFamNet's performance in Appendix Table~\ref{table_detailed_5_cs_results_from_best_acfamnet}, some validation results are better than the training results, which may also be attributed to the small dataset size. The prediction results for two example validation images, illustrated in Appendix Figures~\ref{fig:ACFamNetpro_image_66_cropped} and~\ref{fig:ACFamNetpro_image_122_cropped}, further demonstrate that ACFamNet Pro can accurately count small and clustered colonies. In both cases, the predicted counts (99.94 vs 94 and 147.34 vs 142) are close to the ground truth, and the predicted density map aligns well with colony locations in the input image.

\subsubsection{Ablation studies}
\paragraph{Setup}
An ablation study is conducted with the Synoptics Dataset to analyse the effectiveness of different ACFamNet Pro components. The components are RoI Align, the residual similarity map, and the learnable backbone, which represent the distinctive design elements of ACFamNet Pro. When ACFamNet Pro includes the RoI Align component, its RoI Align output size is~3$\times$3, and otherwise it is replaced with~3$\times$3~RoI pooling. Removing the residual similarity map means the similarity map $\bm{R}$ and the leftmost column of Figure~\ref{fig:acfamnetpro_regression_module} are excluded from the regression module. Likewise, excluding the learnable backbone means the backbone weights are frozen during training and evaluation. The training and evaluation methods are identical to those used in \S~\ref{subsec_para_tuning_acfamnetpro}.

\paragraph{Results}
\begin{table*}[t]
\centering
\caption{Analysis of the effectiveness of different ACFamNet Pro components.}
\label{tab_acfamnet_pro_ablation_study}
\begin{tabular}{>{\centering}p{0.27\textwidth}>{\centering}p{0.05\textwidth}>{\centering}p{0.05\textwidth}>{\centering}p{0.05\textwidth}>{\centering}p{0.05\textwidth}>{\centering}p{0.05\textwidth}>{\centering}p{0.05\textwidth}>{\centering}p{0.05\textwidth}>{\centering\arraybackslash}p{0.05\textwidth}}
    \hline
        Component                       & \multicolumn{8}{c}{Combination}                                                                                            \\ \hline
        RoI Align                        & \texttimes                       & \checkmark & \texttimes & \texttimes & \checkmark & \checkmark & \texttimes & \checkmark \\
        Residual similarity              & \texttimes                       & \texttimes & \checkmark & \texttimes & \checkmark & \texttimes & \checkmark & \checkmark \\
        Learnable backbone               & \texttimes                       & \texttimes & \texttimes & \checkmark & \texttimes & \checkmark & \checkmark & \checkmark \\ \hline
        \multirow{2}{*}{Valid MNAE (\%)} & 11.91                            & 11.16      & 12.18      & 12.14      & 10.85      & 10.62      & 10.76      & \bf{9.62}       \\
                                         & $\pm$1.80                        & $\pm$2.09  & $\pm$3.12  & $\pm$2.39  & $\pm$1.86  & $\pm$2.73  & $\pm$3.35  & \bf{$\pm$3.35}  \\ \hline
\end{tabular}
\end{table*}

The ablation study results in Table~\ref{tab_acfamnet_pro_ablation_study} highlight the critical importance of RoI Align, as the validation MNAE consistently decreases when it is included. For example, the combination of RoI Align with residual similarity and the combination of RoI Align with a learnable backbone perform better than those without RoI Align. In other words, incorporating residual similarity or a learnable backbone individually without RoI Align leads to performance degradation. This suggests that ACFamNet Pro must leverage a learnable backbone to fully benefit from the residual similarity map, and RoI Align plays a key enabling role in this process. The synergy of RoI Align, residual similarity map, and learnable backbone improves ACFamNet Pro from 11.91\% to 9.62\% in validation MNAE.

\subsubsection{Comparison with SAFECount}
\paragraph{Setup}
It is necessary to compare ACFamNet Pro with SAFECount since the former is inspired by the latter. The vanilla SAFECount uses the frozen top three blocks of ResNet-18~\citep{He2015} to extract features. It also uses 3 scale factors and RoI pooling. In this study, the first two frozen blocks of ResNet-18 are used to extract features because images in the Synoptics Dataset are too small to support deeper layers. Likewise, the three upsampling layers in the vanilla SAFECount are reduced to two to accommodate smaller images. Additionally, SAFECount is fine-tuned with RoI Align since it is proven effective in ACFamNet and ACFamNet Pro. The data, training method, and evaluation method are the same to those introduced in \S~\ref{subsec_acfamnet_pro_training}.

\paragraph{Results}

\begin{table}[h!]
\centering
\caption{Comparison between ACFamNet Pro and SAFECount.}
\label{tab_acfamnetpro_vs_safecount_tuned_safecount}
\begin{tabular}{cc}
        \hline

        Model                & Validation MNAE (\%)      \\ \hline
        ACFamNet Pro         & \bf{9.62} $\pm$ \bf{3.35} \\
        Vanilla SAFECount    & 9.86 $\pm$ 1.61           \\
        Fine-tuned SAFECount & 9.79 $\pm$ 2.11           \\ \hline
\end{tabular}
\end{table}

The vanilla SAFECount achieves a mean validation MNAE of 9.86\% with a standard deviation of 1.61\% based on $5$-fold cross-validation, which further improves to 9.79\% with RoI Align, as shown in Appendix Table~\ref{tab_safecount_tuning_RoI_operation}. The performance gain introduced by RoI Align is consistent with the findings from ACFamNet and ACFamNet Pro. The comparison between ACFamNet Pro and SAFECount in Table~\ref{tab_acfamnetpro_vs_safecount_tuned_safecount} shows that ACFamNet Pro outperforms the vanilla SAFECount and the fine-tuned SAFECount by 0.24\% and 0.17\% in validation MNAE, respectively.

\subsubsection{Comparison with other counting methods}
\label{subsec_comparision_of_acfamentpro_vs_others}
\paragraph{Setup}
ACFamNet Pro is further compared with the traditional counting methods introduced in \S~\ref{subsec_compa_acfamnet_vs_traditional_methods}. Similarly, the optimised ACFamNet Pro (3$\times$3~RoI Align, learnable backbone, and 3 scale factors) is trained on the Synoptics training set and evaluated on its test set using a hold-out evaluation. The training configuration follows the same setup introduced in \S~\ref{subsec_para_tuning_acfamnetpro}. For a fair comparison, the same hold-out training strategy is also applied to train the vanilla SAFECount on the Synoptics Dataset.

\paragraph{Results}

\begin{table*}[t]
\centering
\caption{Comparison between ACFamNet Pro and other counting methods.}
\label{tab_acfamnetpro_vs_opencfu_autocellseg}
\begin{tabular}{>{\centering}p{0.16\textwidth}>{\centering}p{0.11\textwidth}>{\centering}p{0.14\textwidth}>{\centering}p{0.12\textwidth}>{\centering}p{0.10\textwidth}>{\centering\arraybackslash}p{0.17\textwidth}}
        \hline

        Metric    & OpenCFU & AutoCellSeg & ACFamNet & ACFamNet Pro & Vanilla SAFECount \\ \hline
        MAE       & 41.12   & 60.92       & 11.54    & 8.88         & 10.91         \\
        RMSE      & 47.76   & 69.87       & 15.56    & 11.66        & 14.64        \\
        MNAE (\%) & 46.57   & 68.73       & 12.52    & \bf{11.25}   & 13.73     \\ \hline
\end{tabular}
\end{table*}

\begin{figure}[h!]
\centering
\includegraphics[width=0.90\linewidth]{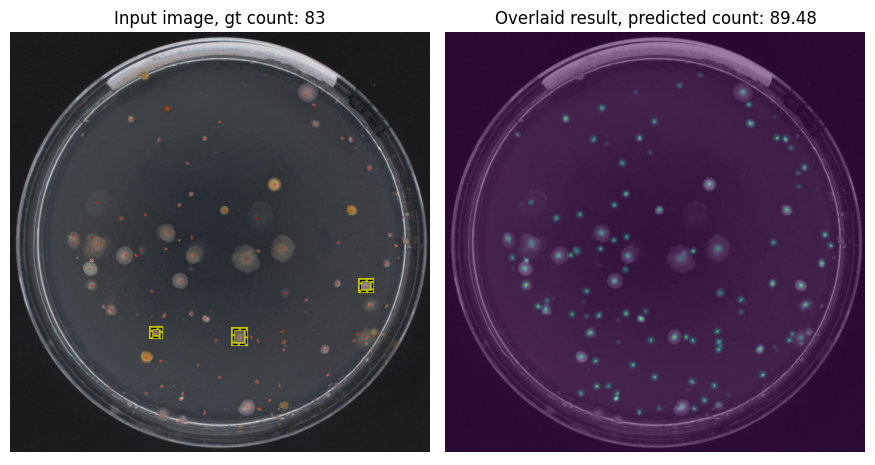}
\caption{Illustration of ACFamNet Pro's prediction. The predicted count and ground truth count are 89.5 and 83, respectively.}
\label{fig:63_from_ACFamNetPro_final}
\end{figure}

ACFamNet Pro achieves a training MNAE of 11.97\% and a test MNAE of 11.25\%, as presented in Appendix Table~\ref{tab_acfamnetpro_and_safecount_final_hold_out_result}. The slightly lower test MNAE compared with the training MNAE may be attributed to the small dataset size. Notably, Table~\ref{tab_acfamnetpro_vs_opencfu_autocellseg} shows that ACFamNet Pro achieves the lowest MNAE at 11.25\%, outperforming the vanilla SAFECount by 2.48\%. This superiority is consistent across the hold-out evaluation results in Table~\ref{tab_acfamnetpro_vs_opencfu_autocellseg} and the $5$-fold cross-validation results in Table \ref{tab_acfamnetpro_vs_safecount_tuned_safecount}. Figure~\ref{fig:63_from_ACFamNetPro_final} illustrates an accurate example prediction (89.48 vs 83) from ACFamNet Pro.

\subsubsection{Category adaptation}
\paragraph{Setup}
Similar to the cross-category evaluation of ACFamNet in \S~\ref{subsec_acfamnet_cross_category_adaptation}, this experiment assesses how well ACFamNet Pro can count small and clustered colonies of a different category. The four plate images shown in Figure~\ref{fig:new_colony_species_two_images} are used to evaluate the trained ACFamNet Pro and SAFECount obtained from \S~\ref{subsec_comparision_of_acfamentpro_vs_others} to avoid repetitive training.

\paragraph{Results}
\begin{table*}[t]
        \centering
        \caption{Comparison of ACFamNet, ACFamNet Pro, and SAFECount on cross-category generalisation.}
        \label{tab_acfamnet_vs_acfamnetpro_vs_safecount_cross_category_prediction}
        \begin{tabular}{ccccc}
                \hline
                Plate image                              & Ground truth & ACFamNet & ACFamNet Pro & SAFECount \\ \hline
                A  & 228          & 306.31   & 211.02       & 124.82       \\
                B & 124          & 475.85   & 285.85       & 78.19       \\
                C  & 529          & 3.1      & 257.14       & 310.79       \\
                D & 302          & 832.12   & 324.28       & 241.98       \\ \hline  \hline

                \multicolumn{2}{c}{MAE}                  & 371.55       & 118.24   & 106.81                      \\ 
                \multicolumn{2}{c}{RMSE}                 & 414.58       & 79.40    & 126.45                      \\ 
                \multicolumn{2}{c}{MNAE(\%)}             & 148.26       & 49.19    & 35.83                      \\ \hline
        \end{tabular}
\end{table*}

Table~\ref{tab_acfamnet_vs_acfamnetpro_vs_safecount_cross_category_prediction} presents the detailed cross-category prediction results of ACFamNet Pro. The MNAE across four plate images shown in Figure~\ref{fig:new_colony_species_two_images} is 49.19\%, indicating ACFamNet Pro retains a degree of generalisation capability when applied to colonies from a different category. Detailed visualisations of these four plate images are provided in Appendix Figures~\ref{fig:2.8_prediction_acfamnetpro}, \ref{fig:envi_image_prediction_acfamnetpro}, \ref{fig:IMG_0431_prediction_acfamnetpro}, and \ref{fig:pour04_prediction_acfamnetpro}. Overall, ACFamNet Pro predicts accurately for Plate Images A and D in Figure~\ref{fig:new_colony_species_two_images}, but performs poorly for Plate Images B and C. One possible reason is that although the multi-head attention mechanism enables ACFamNet Pro to dynamically focus on relevant regions, its attention is still guided by the selected exemplars. This leads to missed colonies that are not represented among the three exemplars, as shown in Appendix Figure~\ref{fig:envi_image_prediction_acfamnetpro}. Additionally, the dramatic colour variation in plate areas between training and evaluation images, as observed in Appendix Figure~\ref{fig:IMG_0431_prediction_acfamnetpro}, may hinder the model's ability to accurately count colonies of a different species.

The comparison of ACFamNet, ACFamNet Pro, and SAFECount on cross-category generalisation is also presented in Table~\ref{tab_acfamnet_vs_acfamnetpro_vs_safecount_cross_category_prediction}, where SAFECount achieves the lowest MNAE of 35.83\%. According to its authors, this is due to the frozen backbone which prevents the model from being overly optimised on the training set, thus improving cross-category generalisation. However, this improved cross-category generalisation comes at the cost of higher errors when counting objects in a similar category as shown in Tables \ref{tab_acfamnetpro_vs_safecount_tuned_safecount} and \ref{tab_acfamnetpro_vs_opencfu_autocellseg}. Additionally, the strong performance of SAFECount on Plate Images B and C appears to contribute significantly to its overall advantage. Furthermore, ACFamNet Pro performs relatively poorly on these two images despite achieving better results on Plate Images A and D.

\subsubsection{Summary}
This section has introduced ACFamNet Pro, which is an improved version of ACFamNet inspired by SAFECount, to count small and clustered colonies. The fine-tuned ACFamNet Pro achieves a mean validation MNAE of 9.62\% with a standard deviation of 3.35\% under $5$-fold cross-validation on the Synoptics training set. The ablation study uncovers that RoI Align consistently improves performance, while the residual similarity map and learnable backbone must operate jointly to yield further gains. The synergy of RoI Align, the residual similarity map, and the learnable backbone reduces the validation MNAE for ACFamNet Pro from 11.91\% to 9.62\% under the same $5$-fold cross-validation. The fine-tuned ACFamNet Pro outperforms vanilla and fine-tuned SAFECount by 0.24\% and 0.17\% in validation MNAE, respectively. Moreover, the model is evaluated on the Synoptics test set after being trained from scratch on the Synoptics training set using a hold-out evaluation. ACFamNet Pro achieves a test MNAE of 11.25\%, outperforming ACFamNet, vanilla SAFECount, OpenCFU, and AutoCellSeg by 1.27\%, 2.48\%, 35.32\%, and 57.48\%, respectively.

This section also has empirically proved that SAFECount can count small bacterial colonies, achieving a mean validation MNAE of 9.86\% and a standard deviation of 1.61\% from $5$-fold cross-validation. ACFamNet Pro further improves counting performance for small and clustered colonies based on limited labelled data. The performance gain is achieved by dynamically weighting objects of interest, optimising gradient flow, and tackling region of interest misalignment. Furthermore, ACFamNet Pro retains a degree of generalisation capability when applied to colonies of a different category, achieving 49.19\% in MNAE across four plate images containing entirely different species. The ablation study reveals that RoI Align is the key component driving ACFamNet Pro's superior performance in counting small and clustered colonies.

\section{Conclusions}
\label{sec-conclusions}
We introduced ACFamNet and ACFamNet Pro to address the research gap that existing counting methods cannot collectively address small object size, object clustering, limited labelled data, and category adaptation. Experimental results demonstrate that both models can effectively learn from limited labelled data to count small and clustered colonies, achieving 12.52\% and 11.25\% in MNAE on the Synoptics held-out test set, respectively. While ACFamNet improves counting
accuracy, it struggles to readily generalise to colonies of a different species. This is likely due to overlapping colony categories within the training images, the limited number of training images, and significant colour variation in plate areas between training and evaluation images.

In contrast, ACFamNet Pro addresses all four challenges by incorporating multi-head attention mechanism and residual connections. It achieves a mean validation MNAE of 9.62\% on the Synoptics training set under $5$-fold cross-validation, outperforming ACFamNet and FamNet by 2.23\% and 12.71\%, respectively. On the Synoptics held-out test set, it further surpasses ACFamNet, SAFECount, OpenCFU, and AutoCellSeg by 1.27\%, 2.48\%, 35.32\%, and 57.48\%, respectively, demonstrating superior generalisation performance. Finally, both models are scalable because in practice the exemplars can be stored in the system and reused for batched input.

This work makes two main contributions. First, we propose ACFamNet, which extends FamNet with an end-to-end trainable model, RoI Align, and an optimised feature extraction module. This design allows it to learn from limited labelled data to count small and clustered colonies. Its performance gains are achieved by addressing region of interest misalignment and improving feature extraction. Second, we propose ACFamNet Pro, which is an advanced version of ACFamNet. It incorporates an additional multi-head attention mechanism and residual connections to collectively address the four challenges identified in the research gap. Compared with ACFamNet, the additional performance gains of ACFamNet Pro are accomplished by dynamically weighting objects of interest and optimising gradient flow.

\section{Limitations and future work}
\label{sec-future-work}
Although the proposed algorithms are rigorously evaluated using both $5$-fold cross-validation and hold-out evaluation, additional statistical analyses in future studies could further validate their improvements. It would also be insightful to compare the proposed few-shot learning-based algorithms with supervised density estimation methods, despite their fundamental differences. The relatively small size of the Synoptics Dataset is acknowledged. Provided by the industry partner Synoptics, it reflects real-world data constraints, and therefore the study emphasises practical system development within this specific context rather than large-scale benchmarking. Additionally, although ACFamNet Pro demonstrates reasonable cross-category generalisation, there remains room for improvement.

Future studies could evaluate ACFamNet Pro on additional datasets, such as AGAR~\citep{majchrowska_agar_2021}, or on larger datasets that contain a wider variety of small and clustered objects. Improving cross-category generalisation remains a key direction. Model pruning, by identifying and removing less critical neurons, could make ACFamNet Pro more compact and efficient. Beyond bacterial colony counting, ACFamNet Pro shows potential in broader applications, including bee-counting~\citep{marstaller_deepbees_2019, rodriguez_recognition_2018}, fly counting~\citep{mamdouh_yolobased_2021, zhong_visionbased_2018}, and corn plant counting~\citep{mota-delfin_detection_2022, wang_convolutional_2021}. Furthermore, it would be valuable to explore if ACFamNet Pro can effectively manage monitoring tasks using satellite imagery, such as counting trees~\citep{abozeid_largescale_2022, yao_tree_2021}, mammals~\citep{xue_automatic_2017, laradji_counting_2020}, and vehicles~\citep{liao_high_2023}.

\section*{CRediT authorship contribution statement}
\textbf{Minghua Zheng:} Conceptualization, Methodology, Software, Formal analysis, Investigation, Data Curation, Writing - Original Draft, Writing - Review \& Editing, Visualization. \textbf{Na Helian:} Resources, Writing - Review \& Editing, Supervision, Project administration, Funding acquisition. \textbf{Peter C. R. Lane:} Resources, Writing - Review \& Editing, Supervision, Funding acquisition. \textbf{Yi Sun:} Resources, Writing - Review \& Editing, Supervision, Funding acquisition. \textbf{Allen Donald:} Resources, Supervision, Project administration, Funding acquisition.

\section*{Declaration of competing interest}
We have nothing to declare.

\section*{Data availability}
The Synoptics Dataset used in this paper is available at github.com/m-zheng/Synoptics-dataset.

\section*{Acknowledgments}
This work was conducted while Minghua Zheng was affiliated with the Department of Computer Science at the University of Hertfordshire. It was partially funded by Hertfordshire Knowledge Exchange Partnership (HKEP), a joint funding provided by European Regional Development Fund (ERDF), University of Hertfordshire, and Synoptics Ltd.


\appendix
\section{Supplementary experimental results for ACFamNet and ACFamNet Pro}
\begin{table}[h!]
\centering
\caption{Detailed $5$-fold cross-validation results of ACFamNet with the best hyper-parameters (k=256,~3$\times$3~RoI Align, and 1 scale factor).}
\label{table_detailed_5_cs_results_from_best_acfamnet}
\renewcommand{\arraystretch}{1.2}
\begin{tabular}{lcccccccc}
        \hline
                                                              & \multirow{2}{*}{Metric} & \multicolumn{5}{c}{Fold} & \multirow{2}{*}{Mean} & \multirow{2}{*}{Std}                                          \\ \cline{3-7}

                                                              &                         & 1                        & 2                     & 3                    & 4     & 5     &            &           \\ \hline
        \multirow{3}{*}{\rotatebox[origin=c]{90}{Training}}   & MAE                     & 16.80                    & 14.89                 & 12.68                & 14.19 & 15.80 & 14.87      & 1.40      \\
                                                              & RMSE                    & 27.79                    & 26.58                 & 23.83                & 25.68 & 27.14 & 26.20      & 1.38      \\
                                                              & MNAE (\%)               & 20.93                    & 15.42                 & 16.24                & 14.21 & 15.05 & 16.37      & 2.37      \\ \cline{1-9}
        \multirow{3}{*}{\rotatebox[origin=c]{90}{Validation}} & MAE                     & 15.42                    & 13.06                 & 19.24                & 5.45  & 7.64  & 12.16      & 5.04      \\
                                                              & RMSE                    & 37.98                    & 19.04                 & 32.09                & 7.42  & 11.29 & 21.56      & 11.76     \\
                                                              & MNAE (\%)               & 13.80                    & 14.03                 & 13.91                & 8.45  & 9.08  & \bf{11.85} & \bf{2.53} \\ \hline
\end{tabular}
\end{table}

\begin{figure}[h!]
\centering
\includegraphics[width=.90\linewidth]{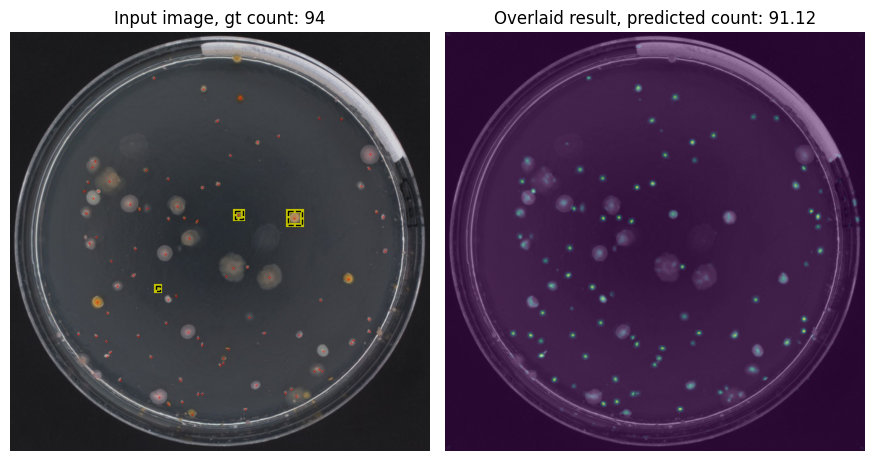}
\caption{ACFamNet's prediction on an unseen image from the validation set. In the left panel, three yellow bounding boxes represent the input exemplars (corresponding to the three blue cubes in Figure~\ref{fig:acfamnet_feature_correlation_module}), and colonies are marked with red dots for illustrative purposes only. The same visualisation scheme is used in the subsequent figures.}
\label{fig:ACFamNet_image_66_cropped}
\end{figure}

\begin{figure}[h!]
\centering
\includegraphics[width=.90\linewidth]{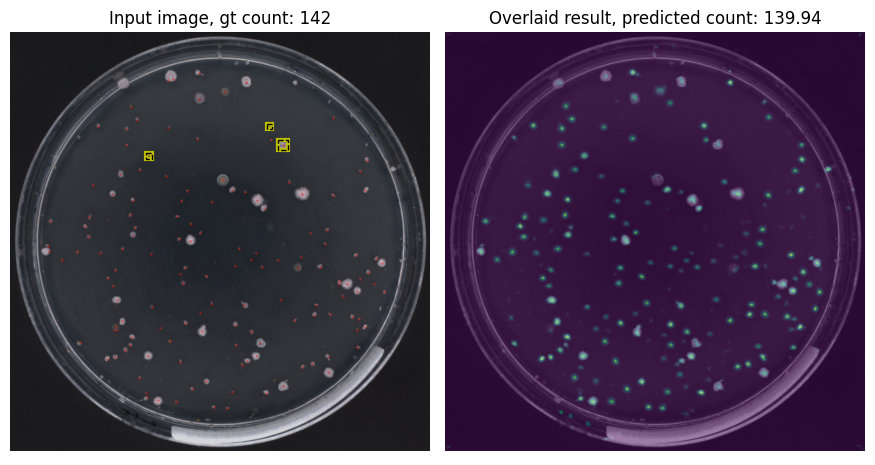}
\caption{ACFamNet's prediction on another unseen image from the validation set.}
\label{fig:ACFamNet_image_122_cropped}
\end{figure}


\begin{table}[h!]
    \centering
    \caption{Results of tuning the number of scale factors for FamNet.}
    \label{tab_famnet_with_diff_scale_factors}
    \begin{tabular}{>{\centering}p{0.1\textwidth}>{\centering}p{0.2\textwidth}>{\centering\arraybackslash}p{0.2\textwidth}}
        \hline
        \multirow{2}{*}{Model}&\multicolumn{2}{c}{Validation MNAE (\%)} \\\cline{2-3}
                    & 3 Scale Factors        & 1 Scale Factor               \\ \hline
        FamNet & \bf{22.33 $\pm$ 6.53} & 23.83 $\pm$ 6.66 \\ \hline
    \end{tabular}
\end{table}

\begin{table}[h!]
        \centering
        \caption{Detailed $5$-fold cross-validation results of the vanilla FamNet (3 scale factors and RoI pooling).}
        \label{table_detailed_5_cs_results_from_best_famnet}
        \renewcommand{\arraystretch}{1.2}
        \begin{tabular}{lcccccccc}
                \hline
                                                                      & \multirow{2}{*}{Metric} & \multicolumn{5}{c}{Fold} & \multirow{2}{*}{Mean} & \multirow{2}{*}{Std}                                          \\ \cline{3-7}

                                                                      &                         & 1                        & 2                     & 3                    & 4     & 5     &            &           \\ \hline
                \multirow{3}{*}{\rotatebox[origin=c]{90}{Training}}   & MAE                     & 16.75                    & 18.39                 & 13.60                & 27.82 & 22.10 & 19.73 & 4.89 \\
                                                                      & RMSE                    & 20.93                    & 27.42                 & 18.14                & 38.30 & 32.11 & 27.38 & 7.33 \\
                                                                      & MNAE (\%)               & 24.86                    & 22.46                 & 19.02                & 27.60 & 22.31 & 23.25 & 2.86 \\ \cline{1-9}
                \multirow{3}{*}{\rotatebox[origin=c]{90}{Validation}} & MAE                     & 28.99                    & 16.41                 & 36.04                & 12.47 & 15.82 & 21.94 & 9.01 \\
                                                                      & RMSE                    & 49.46                    & 25.73                 & 54.00                & 17.64 & 23.07 & 33.98 & 14.79 \\
                                                                      & MNAE (\%)               & 24.68                    & 17.59                 & 32.90                & 13.77 & 22.72 & \bf{22.33} & \bf{6.53} \\ \hline
        \end{tabular}
\end{table}


\begin{table}[h!]
\centering
\caption{Detailed hold-out evaluation results for ACFamNet.}
\label{tab_acfamnet_final_hold_out_result}
\begin{tabular}{ccc}
\hline
    Metric                        & Training set & Test set                                            \\ \hline
    MAE                           & 16.50        & 11.54                             \\
    RMSE                          & 27.54        & 15.56                             \\
    MNAE (\%)                     & 16.64        & \bf{12.52}                              \\ \hline
\end{tabular}
\end{table}


\begin{figure}[h!]
\centering
\includegraphics[width=0.90\linewidth]{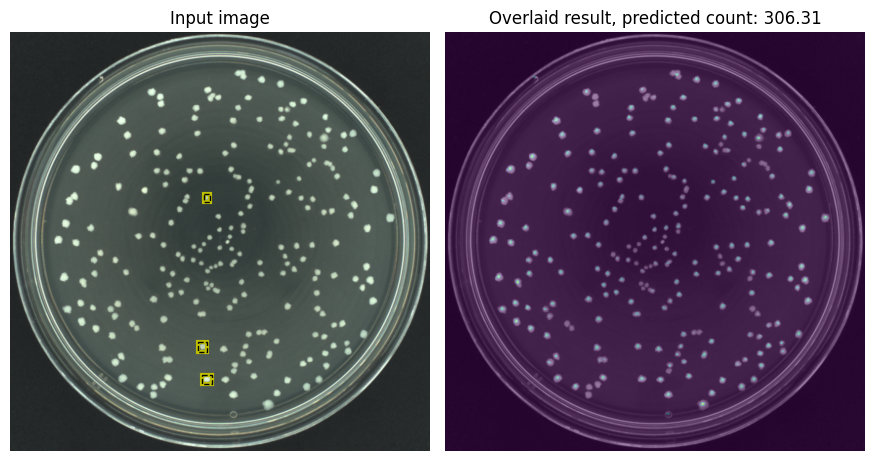}
\caption{Illustration of ACFamNet's prediction on Plate Image A. The predicted count and ground truth count are 306.31 and 228, respectively.}
\label{fig:2.8_prediction_acfamnet}
\end{figure}

\begin{figure}[h!]
\centering
\includegraphics[width=0.90\linewidth]{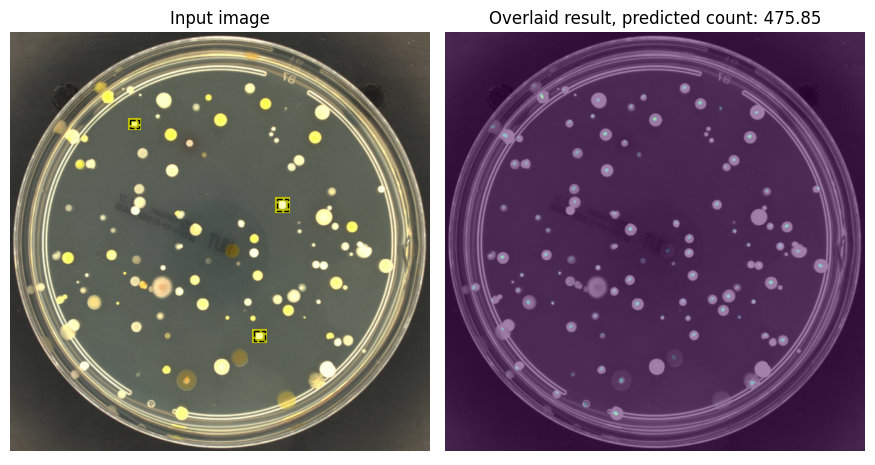}
\caption{Illustration of ACFamNet's prediction on Plate Image B. The predicted count and ground truth count are 475.85 and 124, respectively.}
\label{fig:envi_image_prediction_acfamnet}
\end{figure}

\begin{figure}[h!]
\centering
\includegraphics[width=0.90\linewidth]{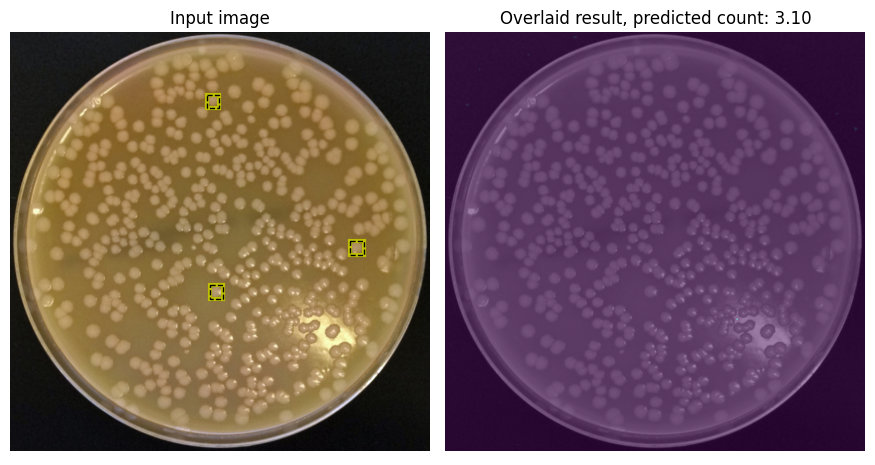}
\caption{Illustration of ACFamNet's prediction on Plate Image C. The predicted count and ground truth count are 3.1 and 529, respectively.}
\label{fig:IMG_0431_prediction_acfamnet}
\end{figure}

\begin{figure}[h!]
\centering
\includegraphics[width=0.90\linewidth]{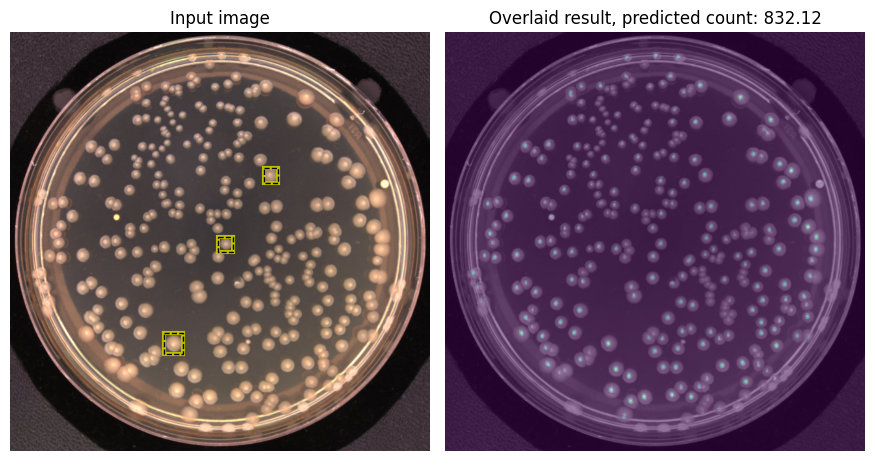}
\caption{Illustration of ACFamNet's prediction on Plate Image D. The predicted count and ground truth count are 832.12 and 302, respectively.}
\label{fig:pour04_prediction_acfamnet}
\end{figure}

\begin{table}[h!]
\centering
\caption{Detailed $5$-fold cross-validation results of ACFamNet Pro with the best hyper-parameters (learnable backbone,~3$\times$3~RoI Align, and 3 scale factors).}
\label{table_detailed_5_cs_results_from_best_acfamnetpro}
\renewcommand{\arraystretch}{1.2}
\begin{tabular}{lcccccccc}
        \hline
                                                              & \multirow{2}{*}{Metric} & \multicolumn{5}{c}{Fold} & \multirow{2}{*}{Mean} & \multirow{2}{*}{Std}                                         \\ \cline{3-7}

                                                              &                         & 1                        & 2                     & 3                    & 4     & 5     &           &           \\ \hline
        \multirow{3}{*}{\rotatebox[origin=c]{90}{Training}}   & MAE                     & 5.95                     & 10.47                 & 11.25                & 7.75  & 7.88  & 8.66      & 1.94      \\
                                                              & RMSE                    & 11.09                    & 17.23                 & 14.06                & 17.49 & 22.48 & 16.47     & 3.81      \\
                                                              & MNAE (\%)               & 8.11                     & 15.97                 & 17.15                & 8.68  & 8.10  & 11.60     & 4.07      \\ \cline{1-9}
        \multirow{3}{*}{\rotatebox[origin=c]{90}{Validation}} & MAE                     & 8.21                     & 9.54                  & 18.21                & 5.06  & 5.96  & 9.40      & 4.68      \\
                                                              & RMSE                    & 15.46                    & 12.65                 & 28.72                & 6.81  & 9.15  & 14.56     & 7.67      \\
                                                              & MNAE (\%)               & 7.41                     & 11.49                 & 15.36                & 7.23  & 6.61  & \bf{9.62} & \bf{3.35} \\ \hline
\end{tabular}
\end{table}

\begin{figure}[h!]
\centering
\includegraphics[width=0.90\linewidth]{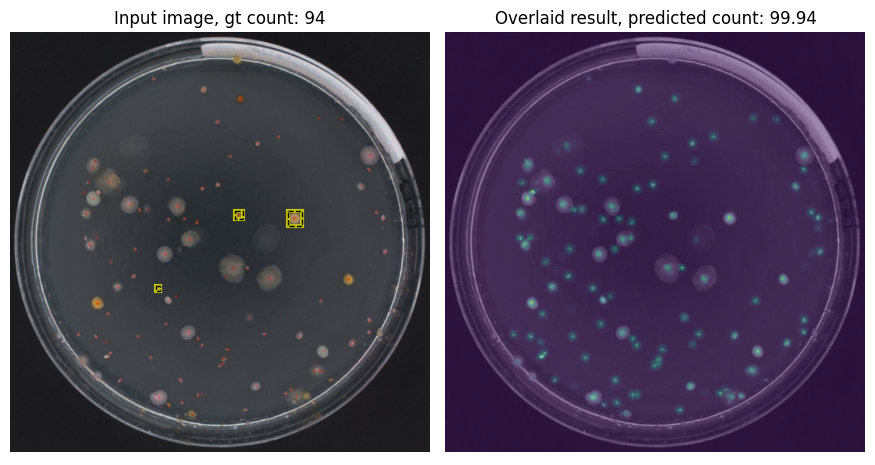}
\caption{ACFamNet Pro's prediction on an unseen image from validation set.}
\label{fig:ACFamNetpro_image_66_cropped}
\end{figure}

\begin{figure}[h!]
\centering
\includegraphics[width=0.90\linewidth]{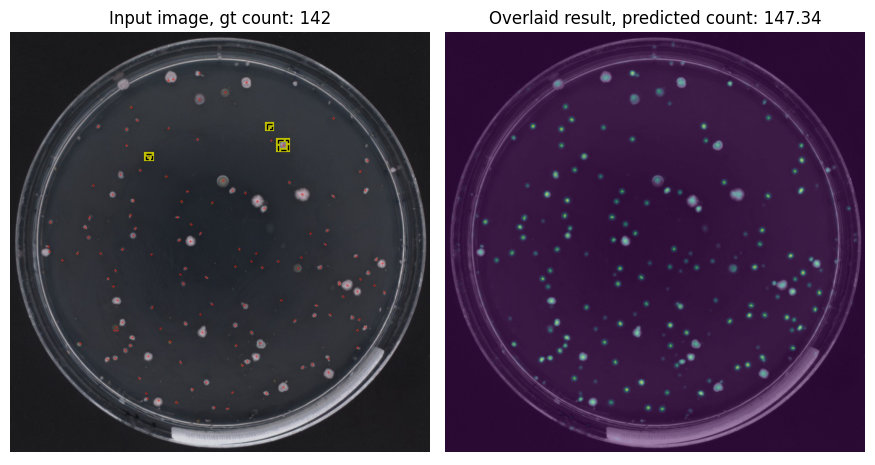}
\caption{ACFamNet Pro's prediction on another unseen image from validation set.}
\label{fig:ACFamNetpro_image_122_cropped}
\end{figure}


\begin{table}[h!]
\centering
\caption{Results of tuning RoI operation for SAFECount.}
\label{tab_safecount_tuning_RoI_operation}
\begin{tabular}{>{\centering}p{0.2\textwidth}>{\centering}p{0.3\textwidth}>{\centering\arraybackslash}p{0.3\textwidth}}
        \hline
        \multirow{2}{*}{Model} & \multicolumn{2}{c}{Validation MNAE (\%)}                        \\\cline{2-3}
                               & 3$\times$3 RoI pooling                    & 3$\times$3 RoI Align  \\ \hline
        SAFECount              & 9.86 $\pm$ 1.61                          & \bf{9.79 $\pm$ 2.11} \\ \hline
\end{tabular}
\end{table}


\begin{table}[h!]
\centering
\caption{Detailed hold-out evaluation results for ACFamNet Pro.}
\label{tab_acfamnetpro_and_safecount_final_hold_out_result}
\begin{tabular}{ccc}
        \hline

        Metric    & Training set & Test set   \\ \hline
        MAE       & 10.38        & 8.88       \\
        RMSE      & 20.34        & 11.66      \\
        MNAE (\%) & 11.97        & \bf{11.25} \\ \hline
\end{tabular}
\end{table}


\begin{figure}[h!]
\centering
\includegraphics[width=.90\linewidth]{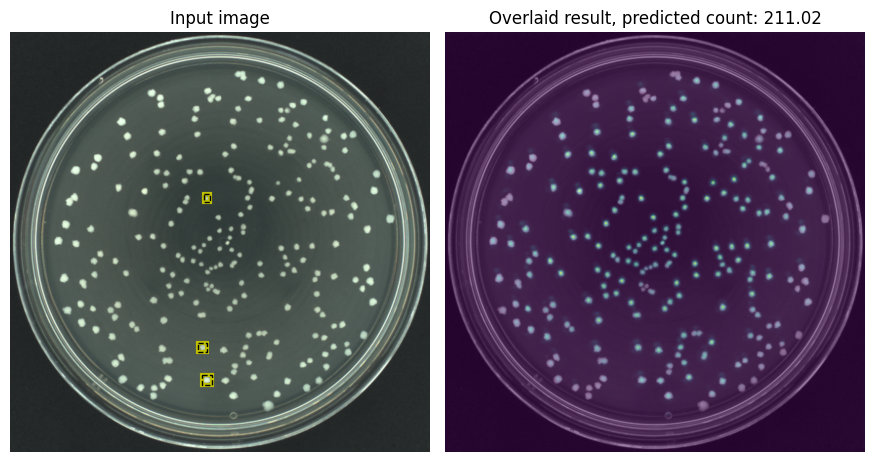}
\caption{Illustration of ACFamNet Pro's prediction on Plate Image A. The predicted count and ground truth count are 211.02 and 228, respectively.}
\label{fig:2.8_prediction_acfamnetpro}
\end{figure}

\begin{figure}[h!]
\centering
\includegraphics[width=.90\linewidth]{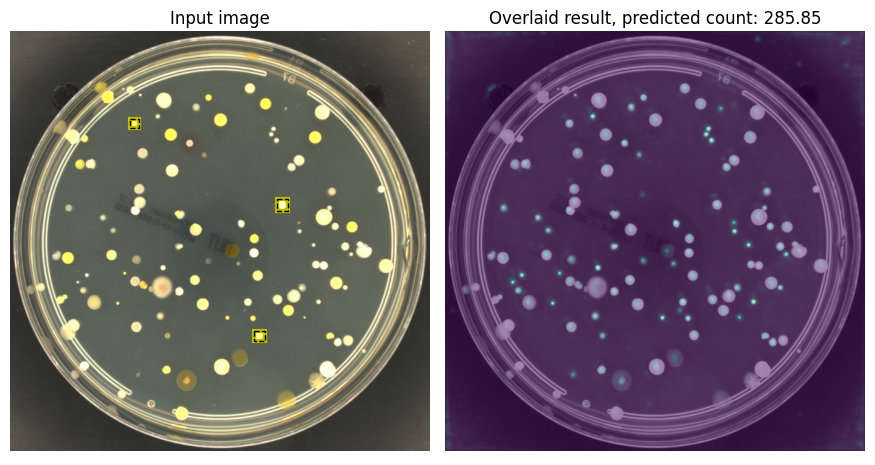}
\caption{Illustration of ACFamNet Pro's prediction on Plate Image B. The predicted count and ground truth count are 285.85 and 124, respectively.}
\label{fig:envi_image_prediction_acfamnetpro}
\end{figure}

\begin{figure}[h!]
\centering
\includegraphics[width=.90\linewidth]{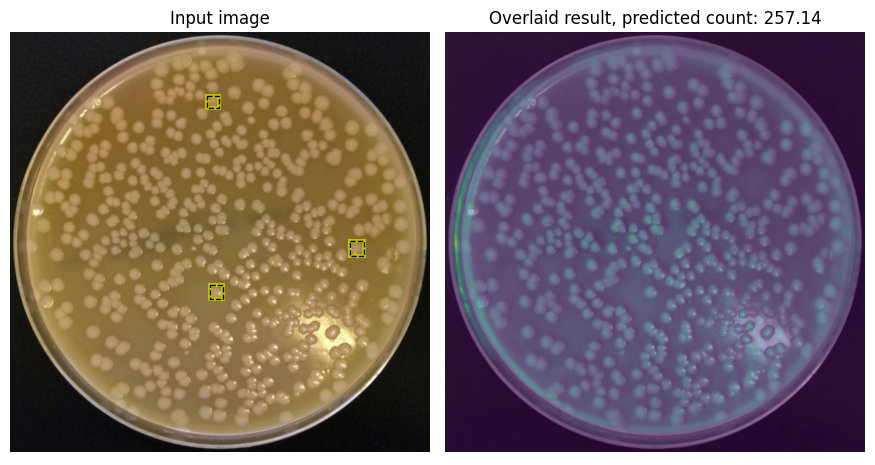}
\caption{Illustration of ACFamNet Pro's prediction on Plate Image C. The predicted count and ground truth count are 257.14 and 529, respectively.}
\label{fig:IMG_0431_prediction_acfamnetpro}
\end{figure}

\begin{figure}[h!]
\centering
\includegraphics[width=.90\linewidth]{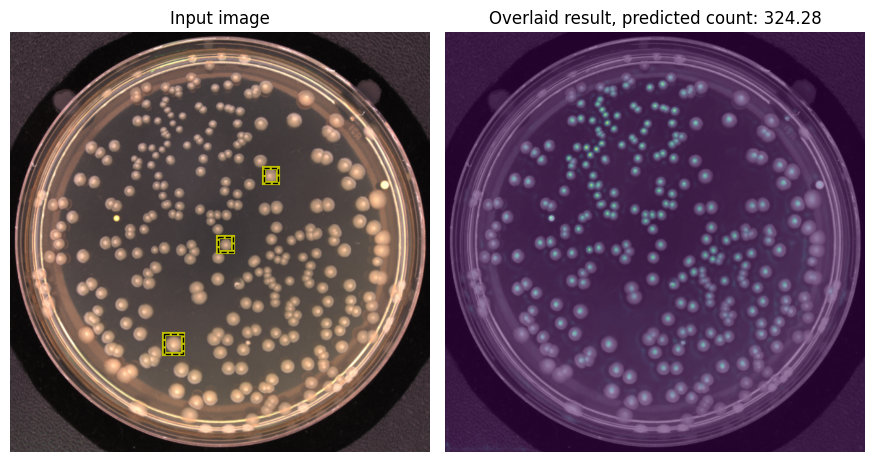}
\caption{Illustration of ACFamNet Pro's prediction on Plate Image D. The predicted count and ground truth count are 324.28 and 302, respectively.}
\label{fig:pour04_prediction_acfamnetpro}
\end{figure}

\clearpage

\bibliographystyle{elsarticle-harv} 
\bibliography{main.bib}

\end{document}